\documentclass[journal,compsoc]{IEEEtran}%10pt,

\usepackage{float}
\usepackage{mathtools}
\usepackage{epsfig}
\usepackage{verbatim}
\usepackage{amssymb}
\usepackage{amsfonts}
\usepackage{dsfont}
\usepackage{amsbsy}
\usepackage{xcolor}
\usepackage{dsfont}
\usepackage{mathbbol}
 
\usepackage{microtype}
\usepackage{graphicx}
\usepackage{subfigure}
\usepackage{booktabs} % for professional tables

\usepackage{hyperref}

\usepackage{algorithm}
\usepackage{algcompatible}
\usepackage{amsmath}
\begin{document}
\title{BiGCN: A Bi-directional Low-Pass\\ Filtering Graph Neural Network}

\author{Zhixian~Chen,
        Tengfei~Ma,
        Zhihua~Jin,
        Yangqiu~Song,
        and~Yang~Wang% <-this % stops a space
\IEEEcompsocitemizethanks{\IEEEcompsocthanksitem Zhixian Chen is with the Department
of Mathematics, Hong Kong University of Science and Technology, Hong Kong SAR,
China.\protect\\
E-mail: zchencz@connect.ust.hk
\IEEEcompsocthanksitem Tengfei Ma is with IBM T. J. Watson Research Center, New York, USA.\protect\\
E-mail: Tengfei.Ma1@ibm.com
\IEEEcompsocthanksitem Zhihua Jin is with the Department
of Computer Science, Hong Kong University of Science and Technology, Hong Kong SAR,
China.\protect\\
E-mail: zjinak@connect.ust.hk

\IEEEcompsocthanksitem Yangqiu Song and Yang Wang are with Hong Kong University of Science and Technology, Hong Kong SAR,
China.\protect\\
E-mail: yqsong@cse.ust.hk, yangwang@ust.hk

}}% <-this % stops an unwanted space}
%\thanks{Manuscript received April 19, 2005; revised August 26, 2015.}}

% The paper headers
\markboth{  }%
{Chen \MakeLowercase{\textit{et al.}}: BiGCN: A Bi-directional Low-Pass Filtering Graph Neural Network}

% The publisher's ID mark at the bottom of the page is less important with
% Computer Society journal papers as those publications place the marks
% outside of the main text columns and, therefore, unlike regular IEEE
% journals, the available text space is not reduced by their presence.
% If you want to put a publisher's ID mark on the page you can do it like
% this:
%\IEEEpubid{0000--0000/00\$00.00~\copyright~2015 IEEE}
% or like this to get the Computer Society new two part style.
%\IEEEpubid{\makebox[\columnwidth]{\hfill 0000--0000/00/\$00.00~\copyright~2015 IEEE}%
%\hspace{\columnsep}\makebox[\columnwidth]{Published by the IEEE Computer Society\hfill}}
% Remember, if you use this you must call \IEEEpubidadjcol in the second
% column for its text to clear the IEEEpubid mark (Computer Society jorunal
% papers don't need this extra clearance.)

% use for special paper notices
%\IEEEspecialpapernotice{(Invited Paper)}

% for Computer Society papers, we must declare the abstract and index terms
% PRIOR to the title within the \IEEEtitleabstractindextext IEEEtran
% command as these need to go into the title area created by \maketitle.
% As a general rule, do not put math, special symbols or citations
% in the abstract or keywords.
\IEEEtitleabstractindextext{%
\begin{abstract}
% Graph convolutional networks have achieved great success on graph-structured data. Many graph-based convolutional networks can be thought of as low-pass filters for graph signals. In this article, we propose a new model \textrm {BiGCN} that represents the graph neural network as a bidirectional low pass filter. Specifically, we not only consider the original graph structure information but also the latent correlation between features, thus \textrm{BiGCN} can filter the signals along with both the original graph and a latent feature-connection graph. Compared with most existing \textrm{GCNs}, \textrm{BiGCN} is more robust and has powerful capacities for denoising. As in real-world networks, noise and mistakes are inevitable in node attributes and connection information, we take feature-noise and structure-mistakes into account. Furthermore, we perform node classification and link prediction in citation networks and co-purchase networks with three settings: \textit{noise-rate}, \textit{noise-level} and \textit{structure-mistakes}. Extensive experimental results demonstrate that our model outperforms the state-of-the-art graph neural networks in both clean and artificially noisy data.

Graph convolutional networks (GCNs) have achieved great success on graph-structured data. Many graph convolutional networks can be thought of as low-pass filters for graph signals. In this paper, we propose a more powerful graph convolutional networks, named \textrm {BiGCN}, that extends to bidirectional filtering. Specifically, we not only consider the original graph structure information but also the latent correlation between features, thus \textrm{BiGCN} can filter the signals along with both the original graph and a latent feature-connection graph. Compared with most existing \textrm{GCNs}, \textrm{BiGCN} is more robust and has powerful capacities for feature denoising. We perform node classification and link prediction in citation networks and co-purchase networks with three settings: \textit{noise-rate}, \textit{noise-level} and \textit{structure-mistakes}. Extensive experimental results demonstrate that our model outperforms the state-of-the-art graph neural networks in both clean and artificially noisy data.

\end{abstract}

% Note that keywords are not normally used for peerreview papers.
\begin{IEEEkeywords}
Noisy graph, low-pass filter, graph convolutional network, node classification, link prediction.
\end{IEEEkeywords}}

% make the title area
\maketitle

% To allow for easy dual compilation without having to reenter the
% abstract/keywords data, the \IEEEtitleabstractindextext text will
% not be used in maketitle, but will appear (i.e., to be "transported")
% here as \IEEEdisplaynontitleabstractindextext when the compsoc 
% or transmag modes are not selected <OR> if conference mode is selected 
% - because all conference papers position the abstract like regular
% papers do.
\IEEEdisplaynontitleabstractindextext
% \IEEEdisplaynontitleabstractindextext has no effect when using
% compsoc or transmag under a non-conference mode.

% For peer review papers, you can put extra information on the cover
% page as needed:
% \ifCLASSOPTIONpeerreview
% \begin{center} \bfseries EDICS Category: 3-BBND \end{center}
% \fi
%
% For peerreview papers, this IEEEtran command inserts a page break and
% creates the second title. It will be ignored for other modes.
\IEEEpeerreviewmaketitle

\IEEEraisesectionheading{\section{Introduction}\label{sec:introduction}}
\IEEEPARstart{G}{raphs} are important research objects in the field of machine learning as they are good carriers for structural data such as social networks, citation networks, and co-purchase networks. Over the past years, there has been a surge of interest in applying deep learning to various graph-based tasks. At the same time, it is well recognized that representation learning methods \cite{cai2018comprehensive,berberidis2019node,rossi2018deep,bahonar2019graph} thoroughly leveraging graph structure and node attributes are fundamental components for the vast majority of graph learning algorithms. In particular, graph convolutional neural networks (\textrm{GCNs}) \cite{kipf2016semi,velivckovic2017graph,hamilton2017inductive,chen2018fastgcn,chang2019local,gao2020topology,bianchi2021graph} received extensive attention due to their impressive performances in various domains. 

\textrm{GCNs} can be interpreted as varied aggregation schemes that propagate node features across graphs. From this point of view, feature noise will spread in the same way, which degrades performance. Unfortunately, in real attributed graphs, feature noise is inevitable. Take social networks as an example. To protect privacy and promote social presence, users can fake or embellish their age, gender, physical characteristics, profession, etc. Thus, the credibility of personal information provided by users is limited. In this paper, we will show how to minimize the impact of noise. Here, we first divide feature noise into two types: \textit{node-wise noise} that exists in nodes and \textit{feature-wise noise} that exists in attributes. The noisy case above is node-wise noise and measurement error is a typical type of feature-wise noise. Below, we state that \textrm {GCNs} can denoise, but fail to cope well with feature-wise noise.

With the success of \textrm{GCNs}, more and more efforts are focused on the reasons why \textrm{GCNs} are so powerful \cite{xu2018powerful}. Li {\em et al} \cite {li2018deeper} re-examined graph convolutional networks (\textrm{GCNs}) and connected it with Laplacian smoothing. NT and Maehara {\em et al} \cite{nt2019revisiting} revisited \textrm{GCNs} in terms of graph signal processing. Interestingly, they found that many graph convolutions can be considered as adjacency induced low-pass filters (e.g.\cite {kipf2016semi, wu2019simplifying}).
 %Several studies (\cite {kipf2016semi, wu2019simplifying}) reported that most spectral-based and some spatial-based graph convolutions act as low-pass filters.
That is, they can capture low-frequency components and remove some high-frequency node-wise noise by making connective nodes more similar. In fact, these findings are not new. Since its first appearance in \cite {bruna2014spectral}, spectral \textrm{GCNs} have been closely related to graph signal processing and denoising.  However, \textrm{GCNs} can't handle all noise as graph filters are row operations for feature matrices while feature-wise noise is column distribution. Furthermore, on isolated nodes or small single components of the graph, their denoising effect is quite limited due to the lack of reliable neighbors.

\begin{figure*}[h!]
\centering 
\includegraphics[width=0.85\textwidth]{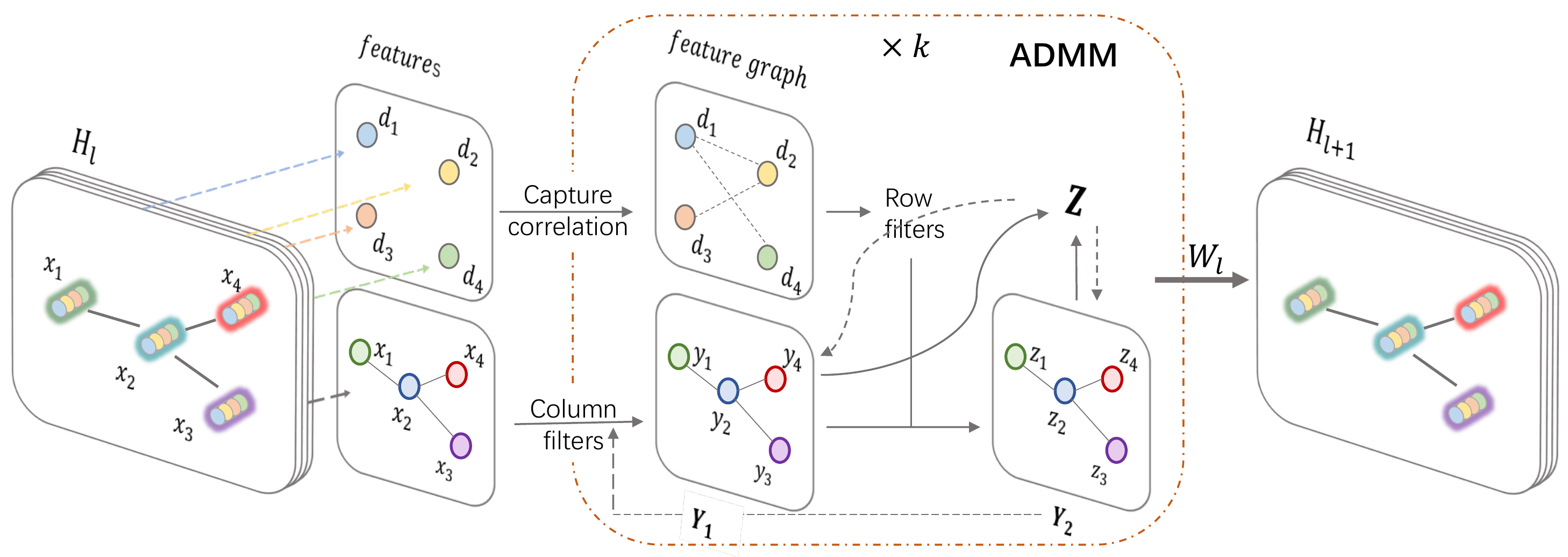}
\caption{Illustration of the $l$-th \textrm{BiGCN} layer. The key innovation of \textrm{BiGCN} is to apply a bi-directional low-pass filter on multi-graph (i.e. the graphs of nodes and features) to incorporate node attributes, node relationships, and feature correlations. We first construct a feature graph: node $d_{i}$ indicates the $i$-th feature dimension with the $i$-th column of $H_l$ as the attribute embeddings; edges are weighted representing given or learnable feature correlations. A bi-directional filter contains a column filter on the node graph and a row filter on the corresponding feature graph. As the input node graph differs from layer to layer, the learned/captured feature correlations, as well as feature graph, are different.
%The latent feature graph is not necessarily dense.
 } 
\label{Fig.model_overview}
\end{figure*}

To address these problems, we design a bi-directional low-pass filter and propose a more powerful graph neural network, called \textrm{BiGCN} (Fig. \ref{Fig.model_overview}). The key point of \textrm{BiGCN} is to introduce a latent \textit{feature graph}. We take feature correlations as weighted edges and represent each feature dimension as a node. If feature correlations are not available, \textrm{BiGCN} can learn to model latent feature correlations. Solving an optimization problem in graph signal processing, we obtain a column filter on the original graph and a row filter on the feature graph to remove node-wise noise and feature-wise noise, respectively.

Finally, we point out that unreliable structure information will limit the denoising capacity of \textrm{GCNs} and make them vulnerable to collapse, while \textrm{BiGCN} can minimize the adverse effect of structural information.  As we mentioned above, graph filters are derived from Graph Fourier Transform and can be formulated as a function with respect to the graph Laplacian. The effect of graph filters depends on eigenvalues and eigenvectors of the Laplacian. However, with incorrect adjacency, most existing spectral graph convolutions are thus no longer the real low-pass filters. Worse, they will block out valid information and introduce noise. In contrast, as \textrm{BiGCN} learned a latent feature correlation graph that is not affected by the given graph structures, it improves the model's fault tolerance to the graph structure mistakes by setting appropriate weights of the two filters. Actually, inaccurate connection information is more common than we ever thought. For instance, on social media, it is difficult to accurately represent social relationships in the real world. Younger users probably have a lot of online friends that users don't know in real life. In addition,  they may be less inclined to follow their acquaintances, such as colleagues or relatives, on social media.

%From the perspective of graph signal processing, our model can extract low-frequency components from the multi-graph, so it is more expressive than the original spectral \textrm{GCNs}.

We evaluate our model on two tasks: node classification and link prediction. In addition to the original graph data, we develop three cases to demonstrate the performance of our model in terms of graph signal denoising and fault tolerance:  1). \textsc{Noise-Rate}: randomly adding Gaussian noise with different variances to a certain percentage of nodes; 2). \textsc{Noise-Level}: adding different levels of Gaussian noise to the whole graph feature; 3). \textsc{Structure-Mistakes}: mistaking a certain percentage of connections. The remarkable performance of our model in extensive experiments confirms our power and robustness in both clean and noisy data.

The main contributions of this work are summarized below.
\begin{itemize}
    \item We propose a new framework for representation learning of attributed graphs. Instead of only considering the signals in the original graph, we take the feature correlations into account and make the model more robust to feature noise as well as structural mistakes.
    \item We formulate our graph neural network based on Laplacian smoothing and derive a bi-directional low-pass graph filter using the Alternating Direction Method of Multipliers (\textrm{ADMM}) algorithm.
    \item We set three cases to demonstrate the powerful denoising capacity and high fault tolerance of our model in tasks of node classification and link prediction.
\end{itemize}

\section{Related Work}

We summarize the related work in the field of graph signal processing and denoising and recent work on spectral graph convolutional networks as follows. 

\subsection{Graph Signal Processing and Denoising}

% Generally, noise is inevitably induced when we use graphs to represent real-world data. First, there will be some mistakes on observed attributions or structure especially when we collect large-scale graph data such as the social network. Second, in most cases, for instance in natural language processing and computer vision tasks, we need to transform text or visual attributes into digital values, i.e. feature vectors, while in the processing of transformation, the noise will be induced and it is hard and expensive to generate feature vectors with perfect expression ability.  In fact, feature vectors themselves are noisy relative to the better representations. Therefore, to enhance the robustness of models and improve performance on graph-based tasks, it is essential to address the negative influence on learning feature representation caused by such unavoidable noise and possible mistakes. 
Graph-structured data is ubiquitous in the world. Graph signal processing (\textrm{GSP})~\cite{ortega2018graph} is intended for analyzing and processing the graph signals whose values are defined on the set of graph vertices. It can be seen as a bridge between classical signal processing and spectral graph theory. One line of the research in this area is the generalization of the Fourier transform to the graph domain and the development of powerful graph filters~\cite{zhu2012approximating,isufi2016autoregressive}. It can be applied to various tasks, such as representation learning and denoising~\cite{chen2014signal}. More recently, the tools of \textrm{GSP} have been successfully used for the definition of spectral graph neural networks, making a strong connection between \textrm{GSP} and deep learning. In this work, we restart with the concepts from graph signal processing and define a new smoothing model for deep graph learning and graph denoising. It is worth mentioning that the concept of denoising/robustness in \textrm{GSP} is different from the defense/robustness against adversarial attacks (e.g. \cite{zugner2019certifiable}), so we do not make comparisons with those models.

% Generally, noise is inevitably induced when we use graphs to represent real-world data. It is common to have mistakes or missing values in node attributes or graph structures. Denoising graph signals have long been studied in the graph signal processing community~\cite{ortega2018graph}.  

\subsection{Spectral Graph Convolutional Networks}
Inspired by the success of convolutional neural networks in images and other Euclidean domains, the researcher also started to extend the power of deep learning to graphs. One of the earliest trends for defining the convolutional operation on graphs is the use of the Graph Fourier Transform and its definition in the spectral domain instead of the original spatial domain~\cite{bruna2014spectral}. Defferrard {\em et al}
\cite{defferrard2016convolutional} proposed ChebyNet which defines a filter as Chebyshev polynomials of the diagonal matrix of eigenvalues, which can be exactly localized in the k-hop neighborhood. Later on, Kipf and Welling~\cite{kipf2016semi} simplified the Chebyshev filters using the first-order polynomial filter, which led to the well-known graph convolutional network. Recently, many new spectral graph filters have been developed. For example, the rational auto-regressive moving average graph filters (\textrm{ARMA})~\cite{isufi2016autoregressive,bianchi2021graph} are proposed to enhance the modeling capacity of GNNs. Compared to the polynomial ones, ARMA filters are more robust and provide a more flexible graph frequency response. Feedback-looped filters~\cite{wijesinghe2019dfnets} further improved localization and computational efficiency. There is also another type of graph convolutional networks that defines convolutional operations in the spatial domain by aggregating information from neighbors. The spatial types are not closely related to our work, so it is beyond the scope of our discussion. As we will discuss later, our model is closely related to spectral graph convolutional networks. We define our graph filter from the perspective of Laplacian smoothing, and then extend it not only to the original graph but also to a latent feature graph in order to improve the capacity and robustness of the model.

% 1. Noisy graph are widely existed in the real word. Some works have proposed some solutions to solve this problem, mitigating the bad effect due to the noise in the graph.

% 2. The paper called Revisiting graph filters proposed a low-pass methods and somehow they conduct a controlled study on how their model work in such noisy cases.  

% 2.1 They propose a theoretical framework to evaluate different graph filters. Why and when those gnn fail to train?

% 2,2 They show that graph convolutional networks is only a low-pass filters. They also demonstrate graph convolutional part doesn't contribute to non-linear manifold learning.

% 3. A new work called diffusion improves graph learning also proposed a method to mitigate the noise in the graph data.

\section{Background: Graph Signal Processing}

In this section, we will briefly introduce some concepts of graph signal processing (\textrm{GSP}), including graphs smoothness, graph Fourier Transform and graph filters, which will be used in later sections. 

\subsection{Graph Laplacian and Smoothness} A graph can be represented as $\mathcal{G}=(\mathcal{V},\mathcal{E})$, which consists of a set of $n$ nodes $\mathcal{V}=\{1,\dots,n\}$ and a set of edges $\mathcal{E} \subseteq \mathcal{V} \times \mathcal{V}$. In this paper, we only consider undirected attributed graphs. We denote the adjacency matrix of $\mathcal{G}$ as $\mathrm{A}=(a_{ij})\in\mathbb{R}^{n\times n}$ and the degree matrix of $\mathcal{G}$ as $\mathrm{D}=diag(d(1),\dots,d(n))\in \mathbb{R}^{n\times n}$. In the degree matrix, $d(i)$ represents the degree of vertex $i\in \mathcal{V}$.  We consider that each vertex $i \in \mathcal{V}$ associates a scalar $x(i) \in \mathbb{R}$ which is also called a graph signal. All graph signals can be represented by $x \in \mathbb{R}^{n}$. Some variants of graph Laplacian can be defined on graph $\mathcal{G}$. We denote the graph Laplacian of $\mathcal{G}$ as $\mathrm{L}=\mathrm{D}-\mathrm{A}\in \mathbb{R}^{n\times n}$.
It should be noted that the sum of rows of graph Laplacian  $\mathrm{L}$ is zero. The smoothness of a graph signal $x$ can be measure through the quadratic form of graph Laplacian: $\Delta(x) = x^\top \mathrm{L}x =  \Sigma_{i,j} \frac{1}{2} a_{ij}(x(i)-x(j))^2$. Due to the fact that $x^\top \mathrm{L}x\ge0$, $\mathrm{L}$ is a semi-positive definite and symmetric matrix.  
% We define variation on node $i$ as $z(i)=\sum_{j \in N_{i}} (x(i)-x(j))$. $N_{i}$ is the one-hop neighborhood of node $i$. It measures the signal variation on node $i$. The variation defined on whole graph can be $\Delta(x) =\sum_{i=1}^{N} z(i) x(i)=\sum_{i=1}^{N} \sum_{j \in N_{i}} (x(i)-x(j)) x(i)=\Sigma_{(i,j)\in E}(x(i)-x(j))^2=x^TLx$.

\subsection{Graph Fourier Transform and Graph Filters} Decomposing the Laplacian matrix with $\mathrm{L} = \mathrm{U} \Lambda \mathrm{U}^\top$, we can get the orthogonal eigenvectors $\mathrm{U}$ as Fourier basis and eigenvalues $\Lambda$ as graph frequencies. The Graph Fourier Transform $\mathcal{F}: \mathbb{R}^{n} \rightarrow \mathbb{R}^{n}$ is defined by $\mathcal{F} x=\hat{x}:=\mathrm{U}^\top x$. The inverse Graph Fourier Transform is defined by $\mathcal{F}^{-1} \hat{x}=x:=\mathrm{U} \hat{x}$. It enables us to transfer the graph signal to the spectral domain, and then define a graph filter $g$ in the spectral domain for filtering the graph signal $x$:
\[
g(\mathrm{L})x = \mathrm{U} g(\Lambda) \mathrm{U}^\top x = \mathrm{U} g(\Lambda) \mathcal{F}(x)
\]
where $g(\Lambda) = diag(g(\lambda_1),...g(\lambda_N))$ controls how the graph frequencies can be altered.

\section{\textrm{BiGCN}}
The Graph Fourier Transform has been successfully used to define various low-pass filters on graph signals (column vectors of feature matrix) and to derive spectral graph convolutional networks~\cite{defferrard2016convolutional,bianchi2021graph,wijesinghe2019dfnets}. A spectral graph convolutional operation can be formulated as a function $g$ concerning the Laplacian matrix $\mathrm{L}$. Although it can smooth the graph and remove certain node-wise noise by assimilating neighbor nodes, it is sensitive to feature-wise noise and unreliable structure information. 
Noting that there are potential correlations between multiple attributes, such information is widely considered in machine learning such as the feature selection process, and can be used to predict one attribute from others. Analogously, we can utilize feature correlations to "denoise" noisy attributes. At the same time, it can reduce the information distortion caused by structure mistakes. In practice, we construct a feature graph based on known or learned feature correlations, and define a new row-directional low-pass filter derived from the Laplacian smoothness assumption on the feature graph. In addition, similar to spectral \textrm{GCNs}, we apply a column-wise filter on the original graph. That is, our model, named \textrm{BiGCN} ( shown in Fig. \ref{Fig.model_overview} ), is an innovative and general spectral \textrm{GCN} with bi-directional low-pass filters. To explain the power of bi-directional graph convolution better, we start with the following simple case.

\subsection{From Laplacian Smoothing to Graph Convolution}
Assuming that $f=y_0+\eta$ is an observation with noise $\eta$, to recover the true graph signal $y_0$, a natural optimization problem is given by: 
\[\min_{y}{\parallel y-f \parallel_2^2+\lambda y^\top \mathrm{L}y},
\]
where $\lambda$ is a hyper-parameter, $\mathrm{L}$ is the (normalized) Laplacian matrix. The optimal solution to this problem is the true graph signal given by
\begin{align}
y=(\mathrm{I}+\lambda \mathrm{L})^{-1}f. 
\label{eq:ori_smooth}
\end{align}

If we generalize the noisy graph signal $f$ to a noisy feature matrix $\mathrm{F}=\mathrm{Y}_0+\mathrm{N}$, then the true graph feature matrix $\mathrm{Y}_0$ can be estimated as follows:
\begin{align}
\mathrm{Y}_0 &= \arg \min_\mathrm{Y}{\parallel \mathrm{Y}-\mathrm{F} \parallel_\mathrm{F}^2+\lambda trace(\mathrm{Y}^\top \mathrm{L}\mathrm{Y})} = (\mathrm{I}+\lambda \mathrm{L})^{-1}\mathrm{F}.
\end{align}
$\mathrm{Y}^\top \mathrm{L}\mathrm{Y}$, the Laplacian regularization, achieves a smoothness assumption on the feature matrix. $(\mathrm{I}+\lambda \mathrm{L})^{-1}$ is equivalent to a low-pass filters in graph spectral domain which can remove node-wise noise and can be used to defined a new graph convolutional operation. Specifically, by multiplying a learnable matrix $\mathrm{W}$ (i.e. adding a linear layer for node feature transformation beforehand, which is similar to \cite{wu2019simplifying,nt2019revisiting}), we obtain a new graph convolutional layer as  follows:
\begin{align}
\mathrm{H}^{(l+1)}=\sigma((\mathrm{I}+\lambda \mathrm{L})^{-1}\mathrm{H}^{(l)}\mathrm{W}^{(l)}).
\label{eq:single_filter}
\end{align}
 In order to reduce the computational complexity, we can simplify the propagation formulation by approximating $(\mathrm{I}+\lambda \mathrm{L})^{-1}$ with its first-order Taylor expansion $\mathrm{I}-\lambda \mathrm{L}$.

\subsection{Bi-directional Smoothing and Filtering}
The key to the success of \textrm{GCNs} is incorporating both node attributes and connection information. While it neglects the relationships between multiple node attributes, which is common in the real world and informative. 

Here, we take the recommendation system as an instance of utilizing feature correlations. The recommendation (such as Netflix) system problem is to predict ratings of movies never seen by users, where rows and columns of rating matrix represent the users and movies, respectively. Additional information like relationships between users (such as their 
interpersonal relations, age, hobbies, education, etc) and movies (such as their genre, director, actors, origin country, etc) can be encoded in the form of a user-graph and a item-graph. This information can be taken advantage of prediction. Users who have lots in common are likely to share the same tastes of movies. On the other hand, users usually have preference toward particular genre/classes of movies and are likely to rate them similarly. 

Inspired by recommendation system problems, we adapt \textrm{GCNs} to multi-graph to incorporate the additional information of feature correlations. We can introduce a "feature adjacency matrix" $\mathrm{A}'$ to indicate this feature connections. In some cases, we can build up a simple adjacency matrix based on rough and limited human knowledge. For instance, let $i$-th, $j$-th, $k$-th dimension feature refer to "height","weight" and "age" respectively, considering that "weight" have very strong correlation with ``height" but weak correlation with ``age", it is reasonable to assign $\mathrm{A}'_{ji} = 1$ while $\mathrm{A}'_{jk} = 0$ (here we assume $\mathrm{A}'\in \{0,1\}^{d\times d}$ for simplification). However, in most cases, it is hard to obtain accurate feature correlations. We endow \textrm{BiGCN} with ability to capture latent relationship between attributes by introducing a learnable adjacency matrix. 

Then we can construct a corresponding "feature graph" $\mathcal{G}_f$ where nodes represent attributes and edges indicate the known/learned feature correlations. Given $\mathrm{Y}\in\mathbb{R}^{n\times d}$, the feature matrix of the original graph $\mathcal{G}$, $\mathrm{Y}^\top$ would be the feature matrix of the feature graph $\mathcal{G}_f$. That is, the rows and columns of $\mathrm{Y}$ are the embeddings of nodes and attributes. 

%These "d feature nodes" form a feature graph with a learnable adjacency matrix $A'_{d\times d}$ and a feature matrix. In addition,  

When noise is not only node-wise but also feature-wise, or when graph structure information is not completely reliable, it is beneficial to consider feature correlation information in order to recover the clean feature matrix better. Thus we add a Laplacian smoothness regularization on feature graph to the optimization problem indicated above:
\begin{align}
\mathcal{L} = &\min_\mathrm{Y} \parallel \mathrm{Y}-\mathrm{F}\parallel_\mathrm{F}^2
\notag\\
&+\lambda_1trace(\mathrm{Y}^\top \mathrm{L}_1\mathrm{Y})+\lambda_2trace(\mathrm{YL}_2\mathrm{Y}^\top).
\end{align}
Here $\mathrm{L}_1$ and $\mathrm{L}_2$ are the normalized Laplacian matrix of the original graph and feature graph, $\lambda_1$ and $\lambda_2$ are hyper-parameters of the two Laplacian regularization. $\mathrm{YL}'\mathrm{Y}^\top$ is the Laplacian regularization on the feature graph or row vectors of the original feature matrix. The solution of this optimization problem is equal to the solution of differential equation:
\begin{align}
\frac {\partial \mathcal{L}}{\partial \mathrm{Y}} &= 2\mathrm{Y}-2\mathrm{F}+2\lambda_1\mathrm{L}_1\mathrm{Y}+2\lambda_2\mathrm{YL}_2=0.
\label{eq:min_condition}
\end{align}

This equation, equivalent to $\lambda_1\mathrm{L}_1\mathrm{Y}+\lambda_2\mathrm{YL}_2 = \mathrm{F}-\mathrm{Y}$, is a Sylvester equation. The numerical solution of Sylvester equations can be calculated using some classical algorithm such as Bartels–Stewart algorithm \cite{bartels1972algorithm}, Hessenberg-Schur method \cite{golub1979hessenberg} and \textrm{LAPACK} algorithm \cite{anderson1999lapack}. However, all of them require Schur decomposition which including Householder transforms and QR iteration with $\mathcal {O}(n^{3})$ computational cost. Consequently, we transform the original problem to a bi-criteria optimization problem with equality constraint instead of solving the Sylvester equation directly:
\begin{align}
\nonumber
&\mathcal{L} =\mathit{\min_{\mathrm{Y}_1}f(\mathrm{Y}_1)+\min_{\mathrm{Y}_2}g(\mathrm{Y}_2)} \;\;\;s.t\;\; \mathrm{Y}_2 -\mathrm{Y}_1 = 0,\\\nonumber
&\mathit{f(\mathrm{Y}_1)} = \frac 12 \parallel \mathrm{Y}_1-\mathrm{F}\parallel_\mathrm{F}^2+\lambda_1\mathit{trace}(\mathrm{Y}_1^\top \mathrm{L}_1\mathrm{Y}_1),\\
&\mathit{g(\mathrm{Y}_2)} = \frac 12 \parallel \mathrm{Y}_2-\mathrm{F}\parallel_\mathrm{F}^2+\lambda_2\mathit{trace}(\mathrm{Y}_2\mathrm{L}_2\mathrm{Y}_2^\top).
\end{align} 

We adopt the \textrm{ADMM} algorithm \cite{boyd2011distributed} to solve this constrain convex optimization problem. The augmented Lagrangian function of $\mathcal{L}$ is:
\begin{align}
\mathcal{L}_p(\mathrm{Y}_1,\mathrm{Y}_2,\mathrm{Z})=&f(\mathrm{Y}_1)+g(\mathrm{Y}_2) \notag \\
&+\mathit{trace}(\mathrm{Z}^\top (\mathrm{Y}_2-\mathrm{Y}_1))+\frac p2\parallel \mathrm{Y}_2-\mathrm{Y}_1\parallel_\mathrm{F}^2.
\end{align}
The update iteration form of \textrm{ADMM} algorithm is:
\begin{align}\nonumber
\mathrm{Y}_1^{(k+1)}&:=\mathit{arg}\min_{\mathrm{Y}_1}\mathcal{L}_p(\mathrm{Y}_1,\mathrm{Y}_2^{(k)},\mathrm{Z}^{(k)})\\\nonumber
&=\mathit{arg}\min_{\mathrm{Y}_1}\frac 12 \parallel \mathrm{Y}_1-\mathrm{F}\parallel_\mathrm{F}^2+\lambda_1\mathit{trace}(\mathrm{Y}_1^\top \mathrm{L}_1\mathrm{Y}_1)\notag \\ &+\mathit{trace}({\mathrm{Z}^{(k)}}^\top (\mathrm{Y}_2^{(k)}-\mathrm{Y}_1))+\frac p2\parallel \mathrm{Y}_2^{(k)}-\mathrm{Y}_1\parallel_\mathrm{F}^2,\\\nonumber
\mathrm{Y}_2^{(k+1)}&:=\mathit{arg}\min_{\mathrm{Y}_2}\mathcal{L}_p(\mathrm{Y}_1^{(k+1)},\mathrm{Y}_2,\mathrm{Z}^{(k)})\\\nonumber
&=\mathit{arg}\min_{\mathrm{Y}_2}\frac 12 \parallel \mathrm{Y}_2-\mathrm{F}\parallel_\mathrm{F}^2+\frac p2\parallel \mathrm{Y}_2-\mathrm{Y}_1^{(k+1)}\parallel_\mathrm{F}^2\notag \\ & +\lambda_2\mathit{trace}(\mathrm{Y}_2\mathrm{L}_2\mathrm{Y}_2^\top)+\mathit{trace}({\mathrm{Z}^{(k)}}^\top(\mathrm{Y}_2-\mathrm{Y}_1^{(k+1)}))  ,\\
\mathrm{Z}^{(k+1)}&=\mathrm{Z}^{(k)}+p(\mathrm{Y}_2^{(k+1)}-\mathrm{Y}_1^{(k+1)}).
\end{align}

We obtain $\mathrm{Y}_1$ and $\mathrm{Y}_2$ iteration formulation by computing the stationary points of $\mathcal{L}_p(\mathrm{Y}_1,\mathrm{Y}_2^{(k)},\mathrm{Z}^{(k)})$ and $\mathcal{L}_p(\mathrm{Y}_1^{(k+1)},\mathrm{Y}_2,\mathrm{Z}^{(k)})$:
\begin{align}\nonumber\label{eq.forward}
\mathrm{Y}_1^{(k+1)}& =\frac 1{1+p}(\mathrm{I}+\frac {2\lambda_1}{1+p}\mathrm{L}_1)^{-1}(\mathrm{F}+p\mathrm{Y}_2^{(k)}+\mathrm{Z}^{(k)}),\\ 
\mathrm{Y}_2^{(k+1)}&=\frac 1{1+p}(\mathrm{F}+p\mathrm{Y}_1^{(k+1)}-\mathrm{Z}^{(k)})(\mathrm{I}+\frac {2\lambda_2}{1+p}\mathrm{L}_2)^{-1}.
\end{align}
    
To decrease the complexity of computation, we can use first-order Taylor approximation to simplify the iteration formulations by choosing appropriate hyper-parameters $p$ and $\lambda_1, \lambda_2$ such that the eigenvalues of $\frac {2\lambda_1}{1+p}\mathrm{L}_1$ and $\frac {2\lambda_2}{1+p}\mathrm{L}_2$ all fall into $[-1,1]$:
\begin{align}\nonumber
\mathrm{Y}_1^{(k+1)}&=\frac 1{1+p}(\mathrm{I}-\frac {2\lambda_1}{1+p}\mathrm{L}_1)(\mathrm{F}+p\mathrm{Y}_2^{(k)}+\mathrm{Z}^{(k)}),\\\nonumber
\mathrm{Y}_2^{(k+1)}&=\frac 1{1+p}(\mathrm{F}+p\mathrm{Y}_1^{(k+1)}-\mathrm{Z}^{(k)})(\mathrm{I}-\frac {2\lambda_2}{1+p}\mathrm{L}_2),\\
\mathrm{Z}^{(k+1)}&=\mathrm{Z}^{(k)}+p(\mathrm{Y}_2^{(k+1)}-\mathrm{Y}_1^{(k+1)}).
\end{align}

In each iteration, as shown in Fig. \ref{Fig.model_overview}, we update $\mathrm{Y}_1$ by appling the column low-pass filter $\mathrm{I}-\frac {2\lambda_1}{1+p}\mathrm{L}_1$ to the previous $\mathrm{Y}_2$, then update $\mathrm{Y}_2$ by appling the row low-pass filter $\mathrm{I}-\frac {2\lambda_2}{1+p}\mathrm{L}_2$ to the new $\mathrm{Y}_1$. To some extent, the new $\mathrm{Y}_1$ is the low-frequency column components of the original $\mathrm{Y}_2$ and the new $\mathrm{Y}_2$ is the low-frequency row components of the new $\mathrm{Y}_1$.
After $k$ iteration (in our experiments, $k=2$), we take the mean of $\mathrm{Y}_1^{(k)}$ and $\mathrm{Y}_2^{(k)}$ as the approximate solution $\mathrm{Y}$, denote it as $\mathrm{Y}=\text{ADMM}(\mathrm{F},\mathrm{L}_1,\mathrm{L}_2)$. In this way, the output of \textrm{ADMM} contains two kinds of low-frequency components. Moreover, we can generalize $\mathrm{L}_2$ to a learnable symmetric matrix based on the original feature matrix $\mathrm{F}$ (or some prior knowledge), since it is hard to give a quantitative description on feature correlations. 

In $l$+1-th propagation layer, $\mathrm{F}=\mathrm{H}^{(l)}$ is the output of $l$-th layer, $\mathrm{L}_2$ is a learnable symmetric matrix depending on $\mathrm{H}^{(l)}$, for this we denote $\mathrm{L}_2$ as $\mathrm{L}_2^{(l)}$. The entire formulation is: 
\begin{align}
\mathrm{H}^{(l+1)} = \sigma (\text{ADMM}(\mathrm{H}^{(l)},\mathrm{L}_1,\mathrm{L}_2^{(l)})\mathrm{W}^{(l)}).
\end{align}

\subsection{Learnable $\mathrm{L}_2$}
 
 We introduce a completely learnable $\mathrm{L}_2$ in our experiments. In detail, we define $\mathrm{L}_2$ as:
 \begin{align*}
     &\mathrm{W}_2 = \textrm{sigmoid}(\mathrm{W}),\\
     &\mathrm{A}_2 = \mathrm{W}_2 + \mathrm{W}_2^\top,\\
     &\mathrm{L}_2= \mathrm{I} -\mathrm{D}_2^{-1/2} \mathrm{A}_2 \mathrm{D}_2^{-1/2},
 \end{align*}
 where $\mathrm{W}$ is an uppertriangle matrix parameter to be optimized. To make it sparse, we also add $L_1$-regularization to $\mathrm{L}_2$. For each layer, $\mathrm{L}_2$ is defined differently. Note that our framework is general and in practice there may be other reasonable choices for $\mathrm{L}_2$.

 \subsection{Discussion about Over-smoothing} Since our algorithm is derived from a bidirectional smoothing, some may worry about the over-smoothing problem. The over-smoothing issue of \textrm{GCN} is explored in \cite{li2018deeper,oono2020graph}, where the main claim is that when the \textrm{GCN} model goes very deep, it will encounter over-smoothing problem and lose its expressive power. From this perspective, our model will also be faced with the same problem when we stack many layers. However, a single \textrm{BiGCN} layer is just a more expressive and robust filter than a normal \textrm{GCN} layer. The general forward function of a single-direction low-pass filtering \textrm{GCN} is $\mathrm{H}^{(l+1)}=\sigma(g(\mathrm{L}_1)\mathrm{H}^{(l)}\mathrm{W}^{(l)})$. Compared with this, $\text{ADMM}(\mathrm{H}^{(l)},\mathrm{L}_1,\mathrm{L}_2^{(l)})$ combines low-frequency components of both column and row vectors of $\mathrm{H}^{(l)}$. It is more informative than $g(\mathrm{L}_1)\mathrm{H}^{(l)}$ since the latter can be regarded as one part of the former to some extent. It also explains that \textrm{BiGCN} is more expressive that single-direction low-pass filtering \textrm{GCNs}. Furthermore, when we take $\mathrm{L}_2$ as an identity matrix (in Equation \ref{eq:min_condition}), \textrm{BiGCN} degenerates to a single-directional \textrm{GCN} with low-pass filter: $((1+\lambda_2)\mathrm{I}+\lambda_1\mathrm{L}_1)^{-1}$. It also illustrates that \textrm{BiGCN} has more general model capacity.
 
 In practice, we can also mix the \textrm{BiGCN} layer with original \textrm{GCN} layers or use jumping knowledge~\cite{xu2018representation} to alleviate the over-smoothing problem: for example, we can use \textrm{BiGCN} at the bottom and then stack other \textrm{GCN} layers above. As we will show in experiments, the adding smoothing term in the \textrm{BiGCN} layers does not lead to over-smoothing; instead, it improves the performance on various datasets. 

\subsection{Model Expressiveness}
%In this section, we add more details about the our discussion of over-smoothing in Section 4. 

% \subsection{First order Approximation of low-pass Filters}
% We reduce the computational complexity of BiGCN by using first order approximation of Taylor expansion which is commonly used in graph neural networks, e.g. APPNP~\citep{klicpera2019predict}. It is a good trade-off between efficiency and accuracy. Especially, in APPNP, they also used a first order approximation to the inverse of a similar matrix to ours and gained very good performance.
% % Note that after the approximation, the bi-directional low-pass filter is different from stacking $L1$ and $L2$ alternatively, because we need a $Z$ to keep $Y_1$ and $Y_2$ close. This is derived from the ADMM, and more theoretically sound than a direct stacking. In fact, stacking $L1$ and $L2$ alternatively is just a special case when we fix $p=0$ in our model (then forward formulation of a BiGCN layer is :
% % \[H^{(l+1)} = \sigma ((1-2\lambda_1L_1)H^{(l)}(1-2\lambda_2L_2)W^{(l)}).
% % \]
% % As shown in Fig \ref{Fig.sub.2}, in most cases non-zero $p$ values (weight of $Z$) make the performance better.  

% \subsection{More Informative Spectral Features}
As a bi-directional low-pass filter, our model can extract more informative features from the spectral domain. To simplify the analysis, let us take just one step of \textrm{ADMM} (k=1). Since $\mathrm{Z}^0 = 0, \mathrm{Y}_1^0 = \mathrm{Y}_2^0 = \mathrm{F}$, we have the final solution from Equation (10) as follows
\begin{align}\nonumber
\mathrm{Y}_1 &=(\mathrm{I}-\frac {2\lambda_1}{1+p}\mathrm{L}_1) \mathrm{F},\\\nonumber
\mathrm{Y}_2 &=(\mathrm{I}-\frac{2 p \lambda_1}{(1+p)^2}\mathrm{L}_1) \mathrm{F} (\mathrm{I}-\frac {2\lambda_2}{1+p}\mathrm{L}_2)\\\nonumber
&= \left((\mathrm{I}-\frac {2\lambda_2}{1+p}\mathrm{L}_2) \mathrm{F}^\top (\mathrm{I}-\frac{2 p \lambda_1}{(1+p)^2}\mathrm{L}_1)\right)^\top.
\end{align}
From this solution, we can see that $\mathrm{Y}_1$ is a low-pass filter which extracts low-frequency features from the original graph via $\mathrm{L}_1$; $\mathrm{Y}_2$ is a low-pass filter which extracts low-frequency features from the feature graph via $\mathrm{L}_2$ and then do some transformation. Since we take the average of $\mathrm{Y}_1$ and $\mathrm{Y}_2$ as the output of $\text{ADMM}(\mathrm{H}, \mathrm{L}_1, \mathrm{L}_2)$, the \textrm{BiGCN} layer will extract low-frequency features from both the graphs. That means, our model adds new information from the latent feature graph while not losing any features in the original graph. Compared to the original single-directional \textrm{GCN}, our model has more informative features and is more powerful in representation. 

When we take more than one step of \textrm{ADMM}, from Equation \ref{eq.forward} we know that the additive component $(\mathrm{I}-\frac {2\lambda_1}{1+p}\mathrm{L}_1) \mathrm{F}$ is always in $\mathrm{Y}_1$ (with a scaling coefficient), and the component $\mathrm{F} (\mathrm{I}-\frac {2\lambda_2}{1+p}\mathrm{L}_2)$ is always in $\mathrm{Y}_2$. So, the output of the \textrm{BiGCN} layer will always contain the low-frequency features from the original graph and the feature graph with some additional features with transformation, which can give us the same conclusion as the one step case.

\section{Experiment}
\label{sec:exp}
We test \textrm{BiGCN} on two graph-based tasks: semi-supervised node classification and link prediction on several benchmarks. As these datasets are usually observed and carefully collected through a rigid screening, noise can be negligible. However, in many real-world data, noise is everywhere and cannot be ignored. To highlight the denoising capacity of the bi-directional filters, we design three cases and conduct extensive experiments on artificial noisy data. In noise-level case, we add different levels of noise to the whole graph. In noise rate case, we randomly add noise to a part of nodes. Considering the potential unreliable connection on the graph, to fully verify the fault tolerance to structure information, we set structure mistakes case in which we will change graph structure. We compare our performance with several baselines including original \textrm{GCN} \cite{kipf2016semi}, \textrm{GraphSAGE} \cite{hamilton2017inductive}, \textrm{GAT} \cite{velivckovic2017graph}, \textrm{GIN} \cite{xu2018powerful}, and \textrm{GDC} \cite{klicpera2019diffusion}.

\subsection{Benchmark Datasets}

\begin{table*}[t]
%\small
\centering
\setlength{\tabcolsep}{2mm}{
\caption{Bechmark Dataset.}
\label{Table.dataset}
\begin{tabular}{lllllll}
  Dataset & Type & Nodes &Edges &Features &Classes &Label Rate 
  \\
  \hline\\
  Cora & Citation & 2,708 &5,278  &1,433 & 7 & 0.052\\
  Citeseer & Citation &3,327   & 4,552 & 3,703 & 6 &0.036 \\
  Pubmed & Citation & 19,717  & 44,324 & 500 &3 &0.003 \\
  DBLP & Citation & 17,716 & 105,734 & 1,639 & 4 & \quad / \\
  AMZ Comp& Co-purchase  & 13,752 & 245,861 & 767 &10 &0.015\\
  AMZ Photos & Co-purchase & 7,650 & 119,081 & 745 & 8&0.021
\end{tabular}}
\end{table*}

We conduct link prediction experiments on Citation networks and node classification experiments both on \textsc{Citation} networks and \textsc{Co-purchase} networks. More statistics details are provided in Table \ref{Table.dataset}.

\begin{itemize}
    \item \textbf{\textsc{Citation}.} 
        A citation network dataset consists of documents as nodes and citation links as directed edges. We use three undirected citation graph datasets: \textsc{Cora} \cite{sen2008collective}, \textsc{CiteSeer} \cite{nr-aaai15}, and \textsc{PubMed} \cite{namata2012query} for both node classification and link prediction tasks as they are common in all baseline approaches. In addition, we add another citation network \textsc{DBLP}~\cite{pang2015optimal} to link prediction tasks.

    \item \textbf{\textsc{Co-purchase}.}
        We also use two Co-purchase networks \textsc{Amazon Computers} \cite{mcauley2015image} and \textsc{Amazon Photos} \cite{shchur2018pitfalls}, which take goods as nodes, to predict the respective product category of goods. The features are bag-of-words node features and the edges represent that two goods are frequently bought together.
\end{itemize}

\subsection{Baseline Models}
%\zhixian{}{We have shown that there are many methods to obtain the adjacency matrix of feature graph or $L_2$ and even we can use another algorithm to solve the Sylvester equation rather than ADMM. In this way, we can define many variants of BiGCN. Across all the experiments, we define a variant of BiGCN which based on ADMM algorithm, and we show that even such a simple version of BiGCN can be powerful.} 
We compare our \textrm{BiGCN} with several state-of-the-art \textrm{GNN} models: 
\begin{itemize}
    \item \textrm{GCN} \cite{kipf2016semi}: A powerful and efficient graph convolutional network for semi-supervised node classification with a first-order approximation of spectral graph convolution. We adapt \textrm{GCN} into link prediction tasks is consistent with the implementation in \textrm{P-GNN} whose implementation of GCN is equivalent to GAE \cite{kipf2016variational}.
    \item \textrm{GraphSAGE} \cite{hamilton2017inductive}: An inductive and spatial variant of graph convolutional networks proposing a general convolution schema and several aggregation methods. In our implementation, we use \textsc{Mean}$(\cdot)$ as the aggregation function.
    \item \textrm{GAT} \cite{velivckovic2017graph}: A graph attention network that assigns learnable weights to known edges based on node attributes. We use a multi-head attention mechanism.
    \item \textrm{GIN} \cite{xu2018powerful}: The graph isomorphism network is as powerful as the Weisfeiler Lehman graph isomorphism test. We remove the graph-level \textsc{Readout}$(\cdot)$ component and use node embeddings learned by \textrm{GIN} directly for node classification and link prediction.
    \item \textrm{GDC} \cite{klicpera2019diffusion}: Graph diffusion convolution based on generalized graph diffusion. We compare one of the variants of \textrm{GDC} which leverages personalized \textrm{PageRank} graph diffusion to improve the original \textrm{GCN}.
\end{itemize}

\subsection{Noise Case}

To highlight the denoising capacity of bi-directional filters, we design the following three cases with artificial noise and mistakes. Cases of noise-level and noise-rate are to add noise to node features while the case of structure-mistakes is to confuse node connections.  

\begin{itemize}
     
    \item  \textsc{Noise-Level case.} In this case, we add different Gaussian noise with zero mean to all the node features in the graph, i.e. to the feature matrix, and use the variance of Gaussian $n_l$ ($n_l \in [0.1,0.9]$) as the quantitative indexes of the noise level. Given the attribute matrix $\mathrm{X}\in\mathbb{R}^{n\times m}$, we generate a noise matrix $\mathrm{N}\in\mathbb{R}^{n\times m}$ whose entities are sampled from $\mathcal{N}(0,n_l^2)$.
    
    \item  \textsc{Noise-Rate case.} In this case, we add Gaussian noise with the random variance to different proportions of nodes, i.e. some rows of the feature matrix, at a random and quantitatively study how the percentage of nodes $n_r$ ( $n_r \in \{0.2,0.4,\dots,1\}$ ) with noisy features impacts the model performances. Given the attribute matrix $\mathrm{X}\in\mathbb{R}^{n\times m}$, we generate a random mask vector $m\in\{0,1\}^n$ with $p(m_i=1)=n_r$ and a random variance vector $v$. Then we generate a noise matrix $\mathrm{N}\in\mathbb{R}^{n\times m}$ in which $\mathrm{N}_{i,\cdot}$ are sampled from $\mathcal{N}(0,(m\odot v)_i^2)$.
    
    \item  \textsc{Structure-Mistakes case.} In practice, it is common and inevitable to observe wrong or interference link information in real-world data, especially in a large-scale network, such as a social network. Therefore, we artificially make random changes with a certain error ratio $r$  in the graph structure, such as removing edges or adding false edges by directly reversing the value of the original adjacency matrix(from 0 to 1 or from 1 to 0) symmetrically to obtain an error adjacency matrix. Given the adjacency matrix $\mathrm{A}\in\mathbb{R}^{n\times n}$, We generate a random matrix $\mathrm{M} \in \{0,1\}^{n\times n}$ with $p(\mathrm{M}_{ij}=0)=r$. To obtain a "noisy" adjacency matrix $\mathrm{\tilde{A}}$, we let $\mathrm{\tilde{A}} = \mathrm{A}\odot \mathrm{\Bar{M}}+(\mathds{1}-\mathrm{A})\odot(\mathds{1}-\mathrm{\Bar{M}})$ with $\mathrm{\Bar{M}}=\mathrm{M}+\mathrm{M}^\top$ and $\mathds{1}$ is a matrix of one. We choose different error ratios for different datasets: $r\in \{1e-5, 2.5e-5, 5e-5, 7.5e-5, 1e-4, 2.5e-4, 5e-4, 7.5e-4, 1e-3\}$ for \textsc{PubMed} and $r\in\{0.001, 0.003, 0.005, 0.007, 0.009, 0.011, 0.013, 0.015\}$ for other datasets.  
\end{itemize}

We conduct all of the above cases on five benchmarks in node classification tasks and the two previous cases on four benchmarks in link prediction tasks.

\subsection{Model Configuration}

%%% TODO, please update it. 
%%% Different settings for node classification and link prediction.
%%% We choose the best models based on early stopping with 100 patience evaluating on the task of node classification while we only train 100 epochs and select the best models based on ten models validation performance recorded each 10 epochs due to the time limits. 
We train a two-layer \textrm{BiGCN} as the same as other baselines using Adam as the optimization method with 0.01 learning rate, $5\times 10^{-4}$ weight decay, and 0.5 dropout rate for all benchmarks and baselines. In the node classification task, we use early stopping with patience 100 to early stop the model training process and select the best performing models based on validation set accuracy. In the link prediction task, we use the maximum 100 epochs to train each classifier and report the test \textrm{ROC-AUC} selected based on the best validation set \textrm{ROC-AUC} every 10 epochs. In addition, we follow the experimental setting from \textrm{P-GNN} (position-aware \textrm{GNN}) and the approach that we adapt \textrm{GCN} into link prediction tasks is consistent with the implementation in \textrm{P-GNN}. We set the random seed for each run and we take mean test results for 10 runs to report the performances.

All the experimental datasets are taken from \textrm{PyTorch} Geometric and we test \textrm{BiGCN} and other baselines on the whole graph while in \textrm{GDC}, only the largest connected component of the graph is selected. Thus, the experimental results we reported of \textrm{GDC} maybe not completely consistent with that reported by \textrm{GDC}. We found that the Citation datasets in \textrm{PyTorch} Geometric are a little different from those used in \textrm{GCN}, \textrm{GraphSAGE}, and \textrm{GAT}. It may be the reason why their accuracy results on Citeseer and Pubmed in node classification tasks are slightly lower than the original papers reported.

All implementations for both node classification and link prediction are based on \textrm{PyTorch} 1.2.0 and \textrm{Pytorch Geometric}\footnote{\url{https://github.com/rusty1s/pytorch_geometric}}. All experiments based on \textrm{PyTorch} are running on one \textrm{NVIDIA} \textrm{GeForce RTX} 2080 \textrm{Ti GPU} using \textrm{CUDA}. The experimental datasets are taken from the \textrm{PyTorch Geometric} platform.

\subsection{Hyper-parameters}
We tune our hyper-parameters for each model using validation data. To accelerate the tedious process of hyper-parameters tuning, we set $\frac{2\lambda_1}{1+p}=\frac{2\lambda_2}{1+p}=\lambda$ and choose different hyper-parameter $p$ for different datasets. For link prediction, we fix all hyper-parameters on all datasets across all experiments: we set $p=8.5$, $\lambda=1.2$, iteration $k=2$, using two layers \textrm{BiGCN} with 32 hidden units and 0.5 dropout trained by 0.01 learning rate. For node classification, we also fix most hyper-parameters on all benchmarks: iteration $k=2$, two layers \textrm{BiGCN} with 16 hidden units and 0.5 dropout trained by 0.01 learning rate. The values of $p$ and $\lambda$ depend on datasets and noise cases. In the case of \textsc{Noise-Rate} and \textsc{noise-level}, we set $p=3$, $\lambda=1.8$ on \textsc{Citation} networks, $p=2.5$, $\lambda=1$ on \textsc{AMZ Computer} and $p=1.5$, $\lambda=0.8$ on \textsc{AMZ Photos} across all the noise settings. In the cases of \textsc{noise-level}, we set $p=0.1$, $\lambda=0.8$ on \textsc{Cora} and \textsc{PubMed}, $p=0.05$, $\lambda=0.8$ on \textsc{Citeseer} and $p=0.1$, $\lambda=1$ on \textsc{co-purchase} Networks across all structural error ratios. 

% For graphs with larger feature dimensions, we can use a precomputed thresholded correlation matrix for $L2$, or we can add the BiGCN layer in the middle of the whole network instead of the bottom, i.e. after the feature dimension is reduced. The sizes of all datasets in our experiments are acceptable and the running time is not increasing too much.

\subsection{Results}

\begin{table*}[h]
\centering 
%\large
\caption{\textrm{BiGCN} compared to \textrm{GNNs} on node classification tasks, measured in accuracy (\%). Standard deviation errors are given.}
\begin{tabular}{cccccc} 
\hline
      & Cora           & Citeseer       & PubMed         & Comp            & Photo           \\
\hline
GCN   & 81.8 $\pm$ 0.6 & 71.0 $\pm$ 0.6 & 78.9 $\pm$ 0.6 & 82.7 $\pm$ 4.6  & 90.8 $\pm$ 1.3  \\
SAGE  & 82.3 $\pm$ 0.5 & 70.5 $\pm$ 0.7 & 78.5 $\pm$ 0.5 & 83.1 $\pm$ 4.2  & 90.8 $\pm$ 1.1  \\ 
GAT   & 83.1 $\pm$ 0.5 & \textbf{71.7} $\pm$ 0.5 & 78.5 $\pm$ 0.5 & 76.3 $\pm$ 3.5 & 88.2 $\pm$ 1.3 \\
GIN   & 79.4 $\pm$ 0.8 & 62.7 $\pm$ 1.2 & 77.7 $\pm$ 0.7   & 41.4 $\pm$ 3.6  & 37.1 $\pm$ 12.0  \\
GDC   & 83.0 $\pm$ 0.6 & 70.7 $\pm$ 0.7 & 77.5  $\pm$ 0.6 & 84.5 $\pm$ 0.8 & 89.7 $\pm$ 0.4 \\
BiGCN & \textbf{83.6} $\pm$ 0.7 & 71.0 $\pm$ 0.6   & \textbf{80.0} $\pm$ 0.3   & \textbf{87.0
} $\pm$ 0.6  & \textbf{92.6} $\pm$ 0.3 \\
\hline
\end{tabular}
\label{Table.node_classification}
\end{table*}

\begin{table*}[h]
\centering 
\caption{\textrm{BiGCN} compared to \textrm{GNNs} on link prediction tasks, measured in \textrm{ROC AUC} (\%). Standard deviation errors are given.}
\begin{tabular}{ccccc}
\hline
      & Cora  & Citeseer & PubMed & DBLP    \\
\hline
GCN   & 89.2 $\pm$ 0.8  & 87.3 $\pm$ 1.7  & 91.7 $\pm$ 0.8   & 92.9 $\pm$ 0.4  \\
SAGE  & 90.4 $\pm$ 0.7  & 89.7 $\pm$ 0.7  & 91.8 $\pm$ 0.3   & 92.6 $\pm$ 0.2    \\
GAT   & 88.6 $\pm$ 0.8  & 87.3 $\pm$ 1.1  & \textbf{92.6}  $\pm$ 0.4& \textbf{93.1} $\pm$ 0.3 \\
GIN   & 87.7 $\pm$ 0.7  & 90.1 $\pm$ 1.3  & 84.7 $\pm$ 0.6  & 91.1 $\pm$ 0.4   \\
GDC   & 89.5 $\pm$ 0.4  & 88.5 $\pm$ 1.1  & 91.6 $\pm$ 0.7  & 92.6 $\pm$ 0.4  \\
BiGCN & \textbf{91.5} $\pm$ 0.5 & \textbf{90.5} $\pm$ 0.7 & 91.6 $\pm$ 0.3  & \textbf{93.1} $\pm$ 0.3  \\
\hline
\end{tabular}
\label{Table.BiGCN_link_prediction}
\end{table*}

\begin{figure*}[!h]
  \centering
    \begin{minipage}[b]{\textwidth}
\includegraphics[width=\linewidth]{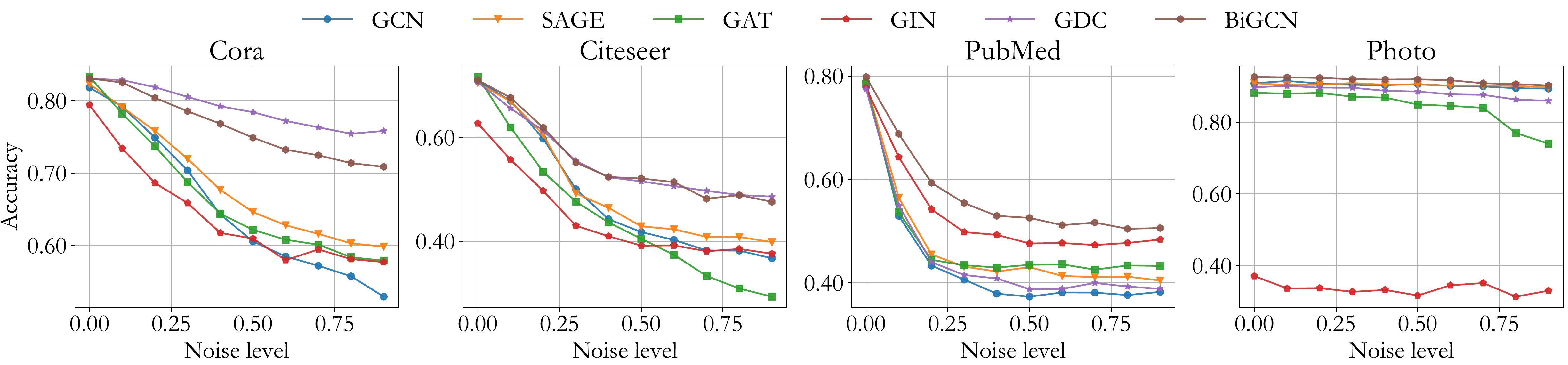}\vspace{5pt}
\includegraphics[width=\linewidth]{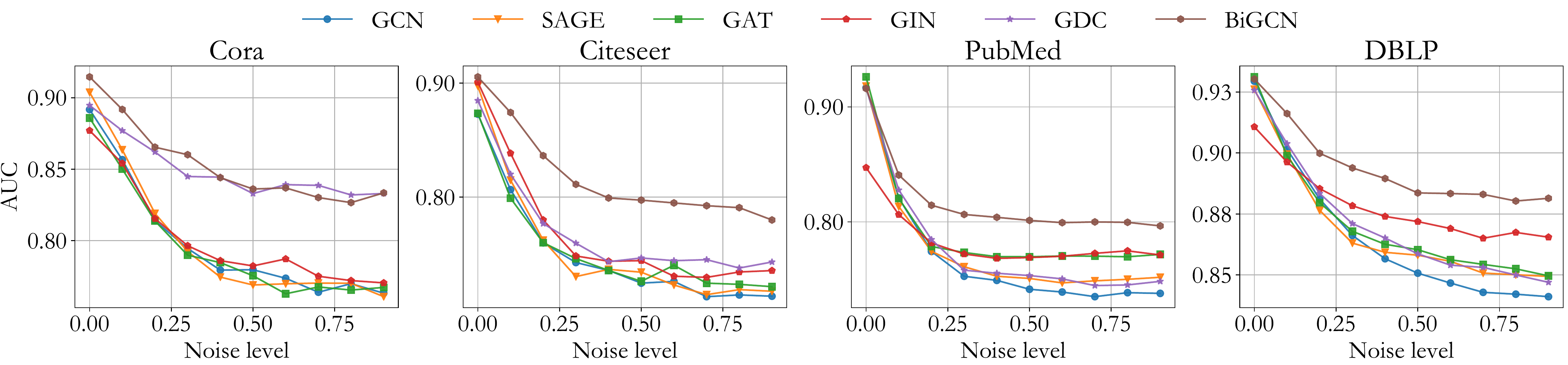}
\end{minipage}
  \caption{In the case of \textsc{Noise-Level}, the accuracy of \textit{Node classification} (upper) and AUC of \textit{link prediction} (lower) of models. For node classification, except \textrm{Cora}, \textrm{BiGCN} performs well on all benchmarks, particularly on \textrm{PubMed}. For link prediction, \textrm{BiGCN} significantly outperform \textrm{GCNs} across all datasets. For instance, it improves AUC score by 4.3\% over the bset baselines (\textrm{GDC}) on \textrm{Citeseer} dataset.}
	\label{fig.noise_level_all_classification_link_prediction}
	\vspace{0.2in}
\end{figure*}

\begin{figure*}[h]
\centering 
\includegraphics[width=0.8\textwidth]{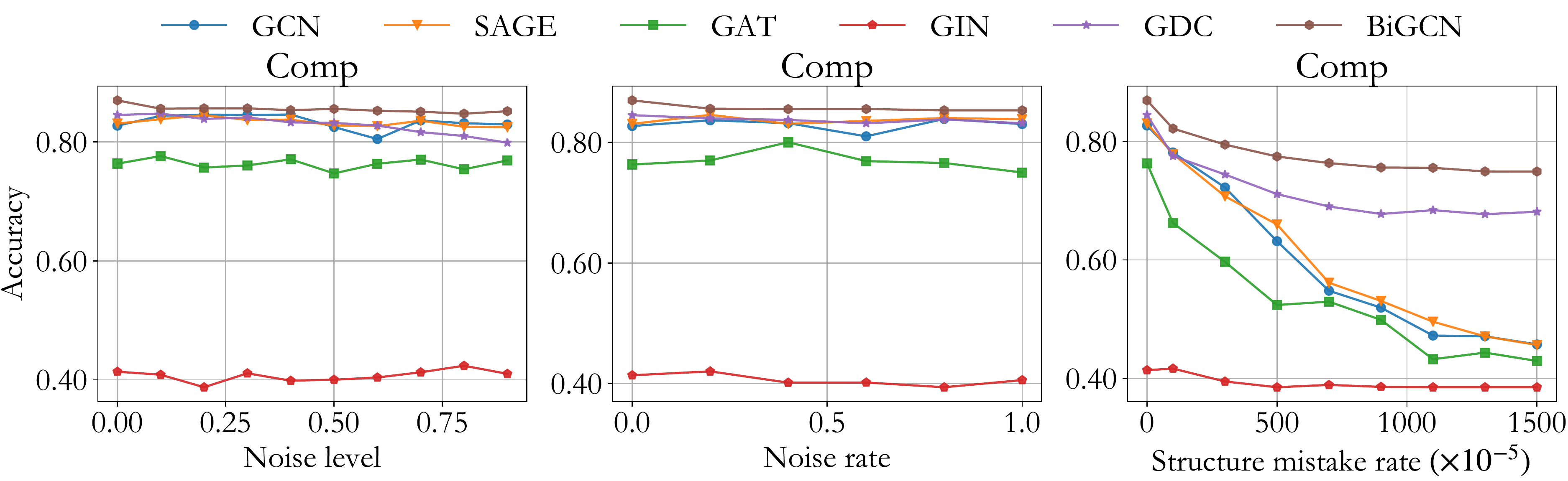}
\caption{Accuracy of \textit{Node classification} of models on \textsc{AMZ Computer} dataset in the cases of \textit{Noise-Level} (right), \textit{Noise-Rate} (middle) and \textit{Structure-Mistakes} (left).  \textrm{BiGCN} outperforms all baselines in all settings and gains 6.7\% improvement in accuracy over the best baseline (\textrm{GDC}).}
\label{Fig.other_noise_comp_node_classification}
\end{figure*}

\begin{figure*}[!h]
  \centering
    \begin{minipage}[b]{\textwidth}
\includegraphics[width=0.95\linewidth]{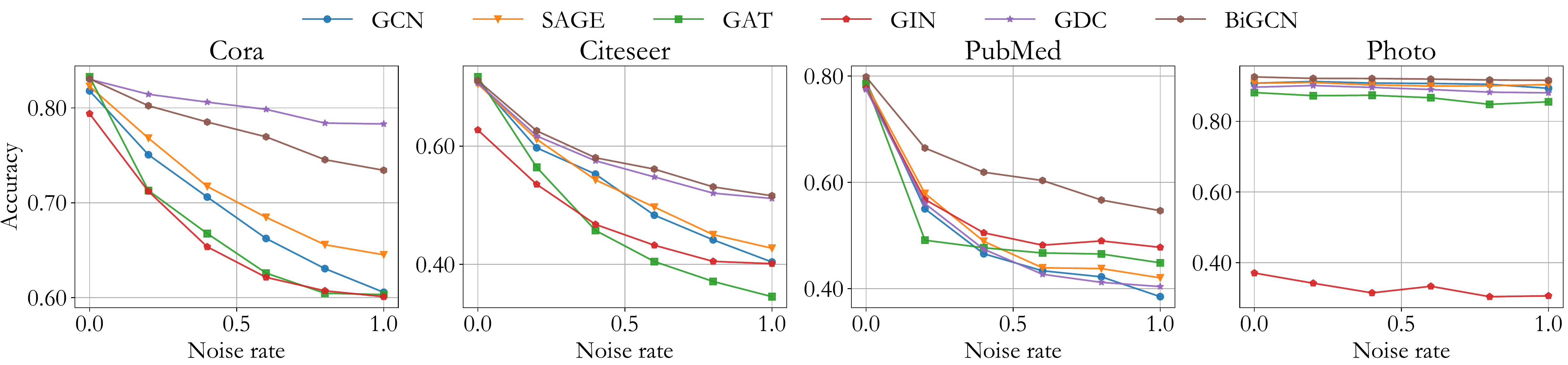}\vspace{4pt}
\includegraphics[width=0.95\linewidth]{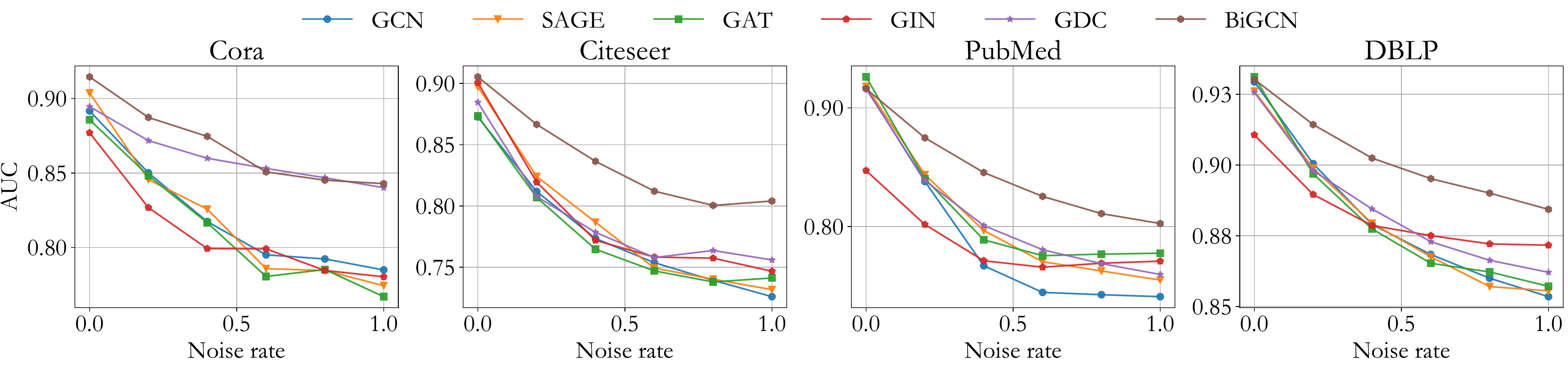}
\end{minipage}
  \caption{In the case of \textsc{Noise-Rate}, accuracy of \textit{node classification} (upper) and AUC of \textit{link prediction} (lower) of models. For node classification, a significant performance gain is observed in \textrm{PubMed} datasets: 4.8\% improvement of \textrm{BiGCN} in the extreme setting over the best baselines (\textrm{GIN}).Forl link prediction, \textrm{BiGCN} also gains the predominant advantage over all baselines in all benchmarks.}
	\label{fig:noise_rate_all_classification_link_prediction}
	\vspace{0.2in}
\end{figure*}

\begin{figure*}[h!]
\centering 
\includegraphics[width=0.95\textwidth]{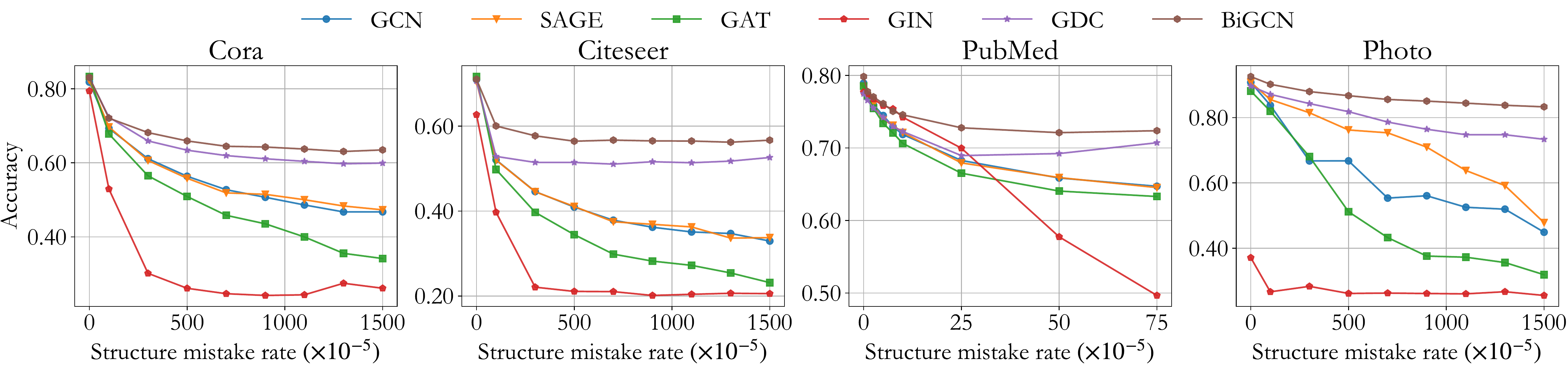}
\caption{In the case of \textsc{Stucture-Mistakes}, accuracy of \textit{node classification} of models. \textrm{BiGCN} shows its impressive robustness and high tolerance to incorrect structure information which makes \textrm{GCNs} collapse.} 
\label{Fig.other_noise_node_classification}
\end{figure*}

We evaluate \textrm{BiGCN} and baselines on node classification and link prediction tasks. To test their denoise capacity and robustness on noisy graphs, we artificially add feature-noise and structural-noise to clean benchmark and set three types of noise cases in terms of noise level, noise rate, and structure mistakes. As far as feature-noise is concerned, we expect our \textrm{BiGCN} to demonstrate its capabilities as graph filters. For structure-noise, we expect the latent feature graph to help correct structural errors in the original graph.

\subsubsection{Clean Data}

The performances of models on clean benchmarks in node classification and link prediction are shown in Table \ref{Table.node_classification} and \ref{Table.BiGCN_link_prediction} respectively. These results correspond to the values with noise level $n_l=0$. On node classification, \textrm{BiGCN} improves accuracy by 0.5\%, 1.1\%, 2.5\% and 1.8\% over the best baselines on \textsc{Cora}, \textsc{PubMed}, \textsc{AMZ Computer} and \textsc{AMZ Photos} respectively. On link prediction, our model outperforms others over all benchmarks except \textsc{PubMed}. 
 
\subsubsection{Noise-level Case}
In this case, we add different noise level to node attributes. Fig. \ref{fig.noise_level_all_classification_link_prediction} and Fig. \ref{Fig.other_noise_comp_node_classification} show results of node classification and link prediction.

On node classification, BiGCN provides the best performance in all datasets except \textsc{Cora} across all noise levels. For co-purchase networks, the curves of all baselines as well as \textrm{BiGCN} are smooth and flat. A possible explanation is that attributes (i.e. product reviews) rarely provide adequate information about product categories. Abnormal pattern is found: GIN behaves worst in \textsc{AMZ Computer} and \textsc{AMZ Photos}. It might be explained in this way: GIN is designed for graph classification task and fails to generalize well to node classification for all datasets. 

In link prediction tasks, BiGCN improves 4.3\%, 3.9\% and 2.8\% \textsc{ROC AUC} than the best baseline on \textsc{Citeseer}, \textsc{PubMed} and \textsc{DBLP}, respectively. It demonstrates our powerful capacity of denoising.

%To sum up, baseline models don't guarantee to consistently behave well when more noise is added into the feature information of nodes. Our model behaves more robust than baselines.
%

\subsubsection{Noise-rate Case}

The results of noise rate case are shown in Fig. \ref{fig:noise_rate_all_classification_link_prediction} and Fig. \ref{Fig.other_noise_comp_node_classification}. 

For node classification, the results are similar as that in noise-level case: \textrm{BiGCN} outperforms most baselines on all datasets across all noise rate with flatter declines. Especially on \textsc{PubMed}, \textrm{BiGCN} improves node classification accuracy by more than 10\%. For link prediction, the performance of \textrm{BiGCN} is at least 0.3\%, 4.8\%, 2.5\% and 1.2\% higher than the best baseline on \textsc{Cora}, \textsc{Citeseer}, \textsc{PubMed} and \textsc{DBLP}, respectively. In this case, BiGCN shows superior robustness than baselines.

%\subsubsection{Adjacency mistakes}
\subsubsection{Structure-mistakes Case} 

Structure mistakes refer to the incorrect interaction relationship among nodes. In this case, we perform node classification whose results are shown in Fig. \ref{Fig.other_noise_node_classification} and Fig. \ref{Fig.other_noise_comp_node_classification}. \textrm{BiGCN} enhances accuracy by 3.6\%, 4.1\%, 1.2\%, 6.7\% and 10.1\% at least on \textsc{Cora}, \textsc{Citeseer}, \textsc{PubMed}, \textsc{AMZ Computer} and \textsc{AMZ Photos}, respectively. Meanwhile, as the number of faulty connections increases, our model shows a much slower decline than baselines, demonstrating our outstanding robustness. The main reason is that our bi-directional filters can effectively utilize information from the latent feature graph and drastically reduce the negative impact of the incorrect structural information.

\subsection{Sensitivity Analysis}

\begin{figure*}[h]
\centering  
%\small
\subfigure[Iteration and $\lambda_2$ analysis.]{
\label{Fig.sub.1}
\includegraphics[width=0.82\textwidth]{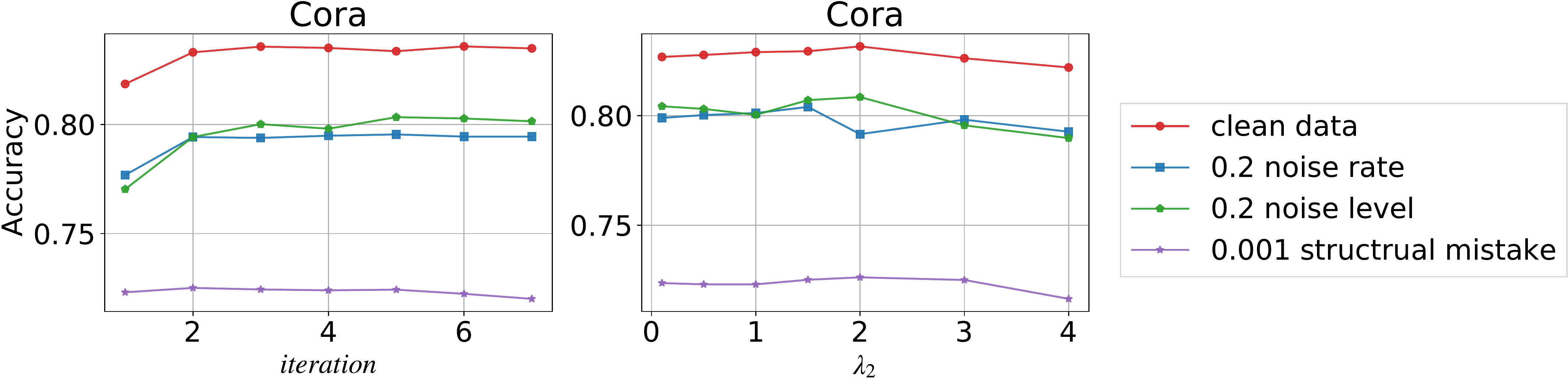}}
\subfigure[$p$ and $\lambda$ analysis on clean data and two cases of noise-rate.]{
\label{Fig.sub.2}
\includegraphics[width=0.95\textwidth]{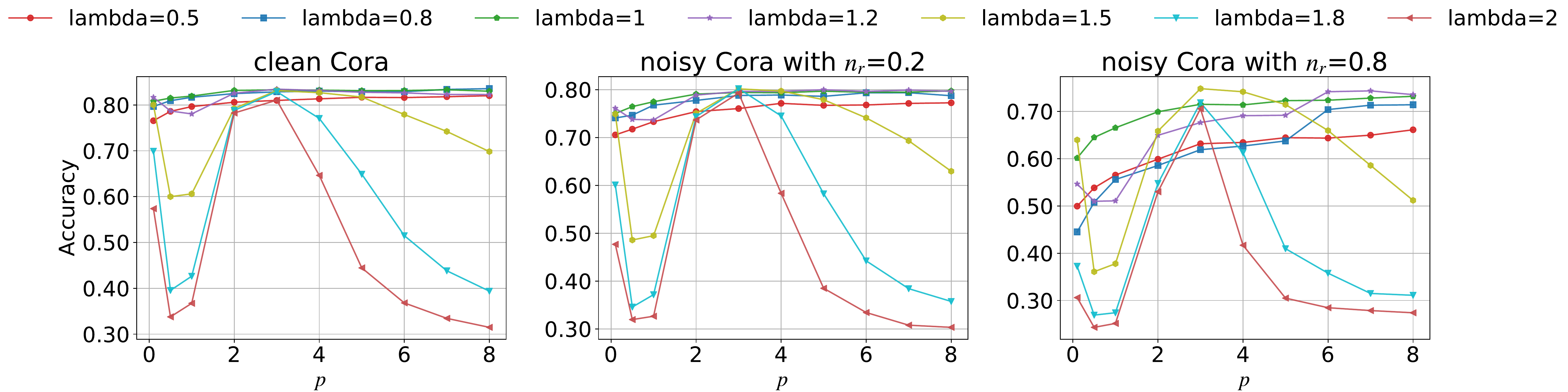}}
\subfigure[$p$ and $\lambda$ analysis on two cases of noise-level and structure-mistakes.]{
\label{Fig.sub.3}
\includegraphics[width=0.95\textwidth]{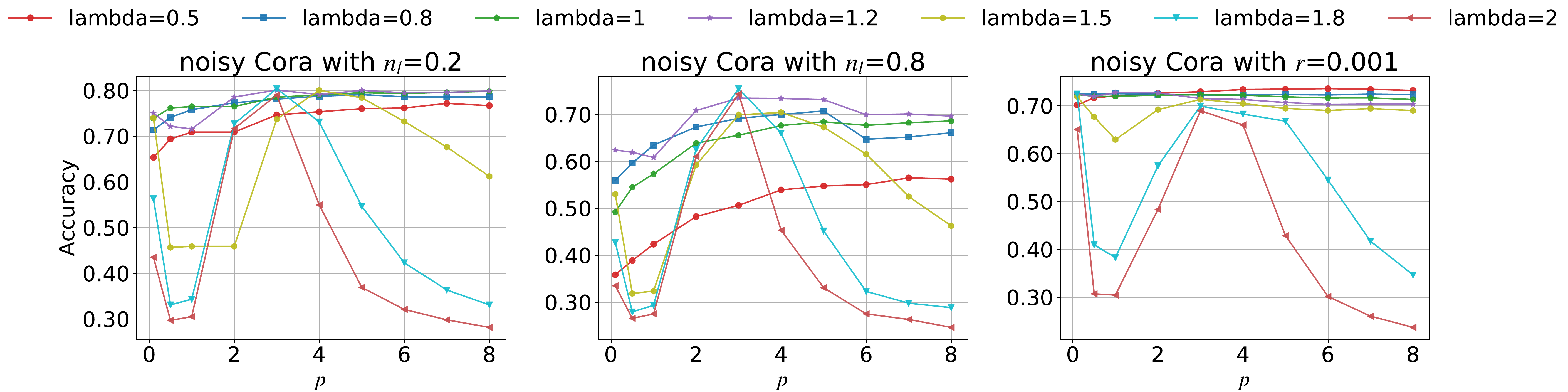}}
\caption{Sensitivity analysis of iteration in \textrm{ADMM}, $\lambda_2$, $\lambda$ and $p$ on \textit{node classification}. For \textit{iteration} ((a) left) and $\lambda_2$ ((a) right), we conduct experiments on clean data and three noise cases with \textit{0.2 noise-rate}, \textit{0.2 noise-level} and \textit{0.1\% structure-mistakes} respectively. For $p$ and $\lambda$, except the mentioned cases, we also provide the performance of \textrm{BiGCN} on Cora with \textit{0.8 noise-rate} ((b) right) and \textit{0.8 noise level} ((c) middle) as well.}
\label{Fig.main}
\end{figure*}

To demonstrate how hyper-parameters (iterations of \textrm{ADMM}, $\lambda_2$, $p$ and $\lambda$) influence BiGCN, we take Cora as an example and present the results on node classification under certain settings of artificial noise.

First, we investigate the influence of iteration and $\lambda_2$ on clean data and three noise cases with $n_r=0.2$ in noise-rate, $n_l=0.2$ noise-level and $r=0.001$ structure-mistakes respectively. Fig. \ref{Fig.sub.1} shows that \textrm{ADMM} with 2 iterations is good enough and the choice of $\lambda_2$ has very little impact on results since it can be absorbed into the learnable $\mathrm{L}_2$.

Then we illustrate how much the performance of \textrm{BiGCN} depends on $p$ and $\lambda$. Experimental results shown in Fig. \ref{Fig.sub.2} and Fig. \ref{Fig.sub.3} demonstrate that: 1). $p$ guarantees relatively stable performance over a wide range values ($p\geq 3$) for all $\lambda \leq 1.2$; 2). When $\lambda \geq 1.5$, things are quite different in the case of $p\geq 3$: accuracy score decreases rapidly as value of $p$ increases. As for the case of $p\leq 2$, erratic fluctuations are observed. However, with a appropriate value of $p$ (e.g. $p=3$), we can still obtain a competitive or even the best performance. 3). $\lambda$ has larger impact in performance on \textsc{Cora} with more noise (0.8 noise-rate and 0.8 noise-level).

\subsection{Flexible Selection of $\mathrm{L}_2$}

\begin{table}[h]
\centering
\caption{Node classification accuracy in noise rate case \\ on Cora dataset of two types of $\mathrm{L}_2$.}
\begin{tabular}{llllll}
Noise rate & 0.200 & 0.400 & 0.600 & 0.800 & 1.000 \\
\hline\\
Fixed $\mathrm{L}_2$   & 0.807 & 0.774 & 0.756 & 0.733 & 0.726 \\
Learnable $\mathrm{L}_2$ & 0.802 & 0.785 & 0.770 & 0.745 & 0.734
\end{tabular}
\label{tab:fixed_L2_1}
\end{table}

\begin{table*}[h]
\centering
\caption{Node classification accuracy in noise-level case on Cora dataset of two types of $\mathrm{L}_2$.}
\begin{tabular}{llllllllll}
Noise level & 0.100 & 0.200 & 0.300 & 0.400 & 0.500 & 0.600 & 0.700 & 0.800 & 0.900 \\
\hline\\
Fixed $\mathrm{L}_2$   & 0.823 & 0.804 & 0.777 & 0.753 & 0.725 & 0.713 & 0.702 & 0.696 & 0.691 \\
Learnable $\mathrm{L}_2$ & 0.825 & 0.804 & 0.785 & 0.768 & 0.749 & 0.732 & 0.725 & 0.714 & 0.709
\end{tabular}
\label{tab:fixed_L2_2}
\end{table*}

\begin{figure*}[h]
\centering  
\subfigure[Node classification.]{
\label{Fig.sub.node}
\includegraphics[width=0.9\textwidth]{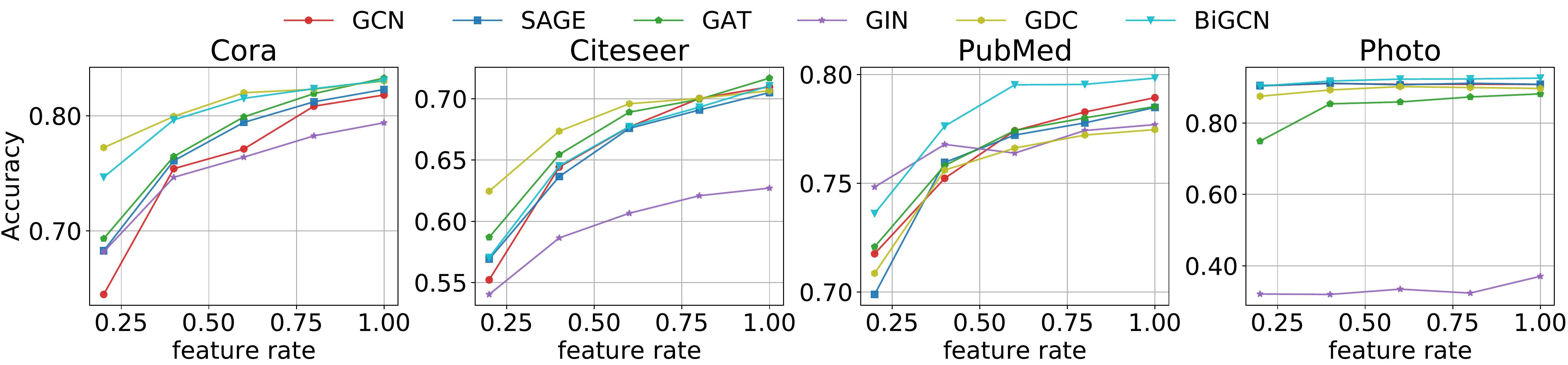}}
\subfigure[Link prediction.]{
\label{Fig.sub.link}
\includegraphics[width=0.9\textwidth]{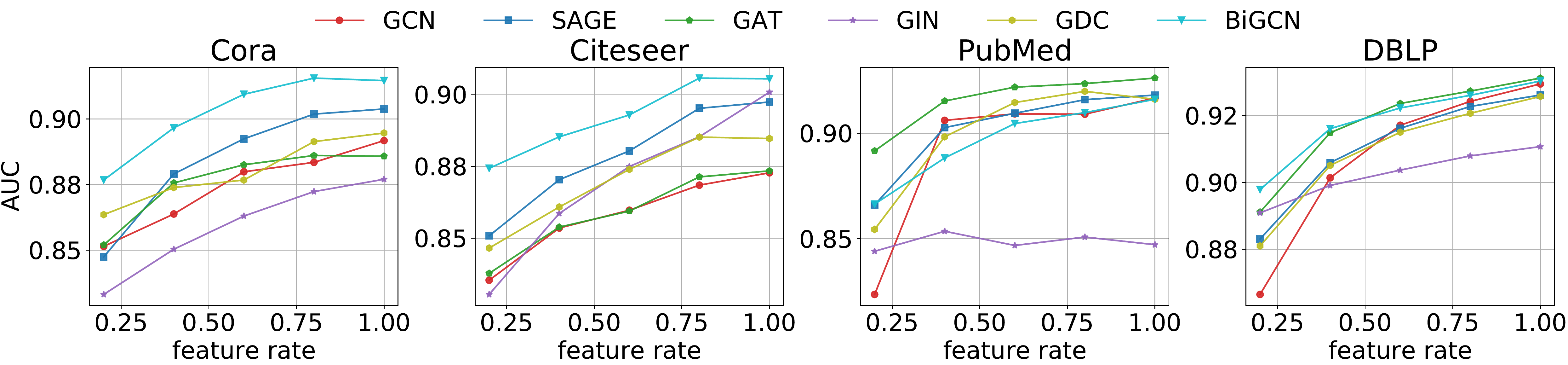}}
\caption{In the case \textsc{Feature-Rate}, accuracy of \textit{node classification} (upper) and AUC of \textit{link prediction} (lower) of models. For node classfication, except \textrm{Citeseer}, \textrm{BiGCN} outperforms \textrm{GCNs} in most settings. For link prediction, significant performance gains in \textrm{Cora} and \textrm{Citeseer} datasets: 1.3\% and 2.4\% improvement of \textrm{BiGCN} over the best baseline (\textrm{GraphSAGE}), respectively.}
\label{Fig.feature-rate}
\end{figure*}

In our paper, we assume the latent feature graph $\mathrm{L}_2$ as a learnable matrix and automatically optimize it. However, in practice it can also be defined as other fixed forms. For example, a common way to deal with the latent correlation is to use a correlation graph~\cite{li2017diffusion}. Another special case is if we define $\mathrm{L}_2$ as an identity matrix, our model will degenerate to a normal (single-directional) low-pass filtering \textrm{GCN}. When we take $\mathrm{L}_2 = I$ in Equation \ref{eq:min_condition}, the solution becomes 
\[
\mathrm{Y} = ((1+\lambda_2)\mathrm{I} + \lambda_1 \mathrm{L}_1)^{-1} \mathrm{F}
\]
which is similar to the single-directional low pass filter (Equation \ref{eq:single_filter}). Then the \textrm{BiGCN} layer will degenerate to the \textrm{GCN} layer as follows:
\[
\mathrm{H}^{(l+1)}=\sigma(((1+\lambda_2)\mathrm{I} + \lambda_1 \mathrm{L}_1)^{-1}\mathrm{H}^{(l)}\mathrm{W}^{(l)}).
\]

To show the difference between different definitions of $\mathrm{L}_2$, we design a simple approach using a thresholded correlation matrix for $\mathrm{L}_2$ to compare with the method used in our main paper. In particular, we define an edge weight $\mathrm{A}_{ij}$ as follows.

\begin{align*}
&(\mathrm{P}_{ij})_{j\in \mathcal{N}(i)\cup {i}} = \text{softmax}([\frac {x_i^\top x_j}{\parallel x_i \parallel \parallel x_j\parallel}]_{j\in \mathcal{N}(i)\cup {i}}),\\
&\mathrm{A}_{ij}=\begin{cases}0,& \mathrm{P}_{ij}\leq \text{mean}(\mathrm{P}) \cr 1, &\mathrm{P}_{ij}>\text{mean}(\mathrm{P})\end{cases}.
\end{align*}
Then we compute $\mathrm{L}_2$ as the normalized Laplacian obtained from $\mathrm{A}$, i.e. $\mathrm{L}_2=\mathrm{\tilde{D}}^{-\frac 12}\mathrm{\tilde{A}}\mathrm{\tilde{D}}^{-\frac 12}$. For a simple demonstration, we only compare the two models on \textsc{Cora} with feature-noise. From Table~\ref{tab:fixed_L2_1} and Table~\ref{tab:fixed_L2_2}, we can see that our learnable $\mathrm{L}_2$ is overall better. However, a fixed $\mathrm{L}_2$ can still give us decent results. When the node feature dimension is large, fixing $\mathrm{L}_2$ may be more efficient.

\subsection{More Discussion and Results in an Additional Case}
Noting that in most benchmarks, node attributes are abundant and even surplus, while in the real world, it is hard to obtain adequate information. In this section, we develop an additional case, the \textsc{Feature-Rate} case, to illustrate that even with limited attributed information, \textrm{BiGCN} can still capture useful and informative feature correlations and improve model performance. In this case, we only keep a portion of feature dimensions (with $r_f$ feature rate, $r_f \in [0.2,0.4,0.6,0.8,1]$), as input of models and perform node classification and link prediction tasks. Extensive experimental results are shown in Fig. \ref{Fig.feature-rate}. It demonstrates that our learned feature adjacency matrix $L_2$ is indeed able to capture effective feature connections making up for the deficiency of attributed information, to some extent.

\section{Conclusion}
We proposed a bi-directional low-pass filtering \textrm{GCN}, a more powerful and robust network than general spectral \textrm{GCNs}. The bi-directional filter of \textrm{BiGCN} can capture more informative graph signal components than the single-directional one. With the help of latent feature correlation, \textrm{BiGCN} also enhances the network's tolerance to noisy graph signals and unreliable edge connections. Extensive experiments show that our model achieves remarkable performance improvement on noisy graphs.

\ifCLASSOPTIONcaptionsoff
  \newpage
\fi

\bibliographystyle{IEEEtran}
% argument is your BibTeX string definitions and bibliography database(s)
\bibliography{main.bib}

% Generated by IEEEtran.bst, version: 1.14 (2015/08/26)
\begin{thebibliography}{10}
\providecommand{\url}[1]{#1}
\csname url@samestyle\endcsname
\providecommand{\newblock}{\relax}
\providecommand{\bibinfo}[2]{#2}
\providecommand{\BIBentrySTDinterwordspacing}{\spaceskip=0pt\relax}
\providecommand{\BIBentryALTinterwordstretchfactor}{4}
\providecommand{\BIBentryALTinterwordspacing}{\spaceskip=\fontdimen2\font plus
\BIBentryALTinterwordstretchfactor\fontdimen3\font minus
  \fontdimen4\font\relax}
\providecommand{\BIBforeignlanguage}[2]{{%
\expandafter\ifx\csname l@#1\endcsname\relax
\typeout{** WARNING: IEEEtran.bst: No hyphenation pattern has been}%
\typeout{** loaded for the language `#1'. Using the pattern for}%
\typeout{** the default language instead.}%
\else
\language=\csname l@#1\endcsname
\fi
#2}}
\providecommand{\BIBdecl}{\relax}
\BIBdecl

\bibitem{cai2018comprehensive}
H.~Cai, V.~W. Zheng, and K.~C.-C. Chang, ``A comprehensive survey of graph
  embedding: Problems, techniques, and applications,'' \emph{IEEE Transactions
  on Knowledge and Data Engineering}, vol.~30, no.~9, pp. 1616--1637, 2018.

\bibitem{berberidis2019node}
D.~Berberidis and G.~B. Giannakis, ``Node embedding with adaptive similarities
  for scalable learning over graphs,'' \emph{IEEE Transactions on Knowledge and
  Data Engineering}, 2019.

\bibitem{rossi2018deep}
R.~A. Rossi, R.~Zhou, and N.~K. Ahmed, ``Deep inductive graph representation
  learning,'' \emph{IEEE Transactions on Knowledge and Data Engineering},
  vol.~32, no.~3, pp. 438--452, 2018.

\bibitem{bahonar2019graph}
H.~Bahonar, A.~Mirzaei, S.~Sadri, and R.~C. Wilson, ``Graph embedding using
  frequency filtering,'' \emph{IEEE Transactions on Pattern Analysis and
  Machine Intelligence}, vol.~43, no.~2, pp. 473--484, 2019.

\bibitem{kipf2016semi}
T.~N. Kipf and M.~Welling, ``Semi-supervised classification with graph
  convolutional networks,'' in \emph{International Conference on Learning
  Representations}, 2017.

\bibitem{velivckovic2017graph}
P.~Veli{\v{c}}kovi{\'c}, G.~Cucurull, A.~Casanova, A.~Romero, P.~Lio, and
  Y.~Bengio, ``Graph attention networks,'' in \emph{International Conference on
  Learning Representations}, 2018.

\bibitem{hamilton2017inductive}
W.~Hamilton, Z.~Ying, and J.~Leskovec, ``Inductive representation learning on
  large graphs,'' in \emph{Advances in Neural Information Processing Systems},
  2017, pp. 1024--1034.

\bibitem{chen2018fastgcn}
J.~Chen, T.~Ma, and C.~Xiao, ``Fastgcn: fast learning with graph convolutional
  networks via importance sampling,'' in \emph{International Conference on
  Learning Representations}, 2018.

\bibitem{chang2019local}
J.~Chang, L.~Wang, G.~Meng, Q.~Zhang, S.~Xiang, and C.~Pan, ``Local-aggregation
  graph networks,'' \emph{IEEE Transactions on Pattern Analysis and Machine
  Intelligence}, vol.~42, no.~11, pp. 2874--2886, 2019.

\bibitem{gao2020topology}
H.~Gao, Y.~Liu, and S.~Ji, ``Topology-aware graph pooling networks,''
  \emph{arXiv preprint arXiv:2010.09834}, 2020.

\bibitem{bianchi2021graph}
F.~M. Bianchi, D.~Grattarola, C.~Alippi, and L.~Livi, ``Graph neural networks
  with convolutional arma filters,'' \emph{arXiv preprint arXiv:1901.01343},
  2019.

\bibitem{xu2018powerful}
K.~Xu, W.~Hu, J.~Leskovec, and S.~Jegelka, ``How powerful are graph neural
  networks?'' in \emph{International Conference on Learning Representations},
  2019.

\bibitem{li2018deeper}
Q.~Li, Z.~Han, and X.-M. Wu, ``Deeper insights into graph convolutional
  networks for semi-supervised learning,'' in \emph{Proceedings of the
  Thirty-Second {AAAI} Conference on Artificial Intelligence}, 2018, pp.
  3538--3545.

\bibitem{nt2019revisiting}
H.~NT and T.~Maehara, ``Revisiting graph neural networks: All we have is
  low-pass filters,'' \emph{arXiv preprint arXiv:1905.09550}, 2019.

\bibitem{wu2019simplifying}
F.~Wu, A.~H.~S. Jr., T.~Zhang, C.~Fifty, T.~Yu, and K.~Q. Weinberger,
  ``Simplifying graph convolutional networks,'' in \emph{Proceedings of the
  36th International Conference on Machine Learning}, vol.~97, 2019, pp.
  6861--6871.

\bibitem{bruna2014spectral}
J.~Bruna, W.~Zaremba, A.~Szlam, and Y.~Lecun, ``Spectral networks and locally
  connected networks on graphs,'' in \emph{International Conference on Learning
  Representations}, 2014.

\bibitem{ortega2018graph}
A.~Ortega, P.~Frossard, J.~Kova{\v{c}}evi{\'c}, J.~M. Moura, and
  P.~Vandergheynst, ``Graph signal processing: Overview, challenges, and
  applications,'' \emph{Proceedings of the IEEE}, vol. 106, no.~5, pp.
  808--828, 2018.

\bibitem{zhu2012approximating}
X.~Zhu and M.~Rabbat, ``Approximating signals supported on graphs,'' in
  \emph{IEEE International Conference on Acoustics, Speech and Signal
  Processing}, 2012, pp. 3921--3924.

\bibitem{isufi2016autoregressive}
E.~Isufi, A.~Loukas, A.~Simonetto, and G.~Leus, ``Autoregressive moving average
  graph filtering,'' \emph{IEEE Transactions on Signal Processing}, vol.~65,
  no.~2, pp. 274--288, 2016.

\bibitem{chen2014signal}
S.~Chen, A.~Sandryhaila, J.~M. Moura, and J.~Kovacevic, ``Signal denoising on
  graphs via graph filtering,'' in \emph{IEEE Global Conference on Signal and
  Information Processing}, 2014, pp. 872--876.

\bibitem{zugner2019certifiable}
D.~Z{\"u}gner and S.~G{\"u}nnemann, ``Certifiable robustness and robust
  training for graph convolutional networks,'' in \emph{Proceedings of the 25th
  ACM SIGKDD International Conference on Knowledge Discovery \& Data Mining},
  2019, pp. 246--256.

\bibitem{defferrard2016convolutional}
M.~Defferrard, X.~Bresson, and P.~Vandergheynst, ``Convolutional neural
  networks on graphs with fast localized spectral filtering,'' in
  \emph{Advances in Neural Information Processing Systems}, 2016, pp.
  3844--3852.

\bibitem{wijesinghe2019dfnets}
W.~A.~S. Wijesinghe and Q.~Wang, ``Dfnets: Spectral cnns for graphs with
  feedback-looped filters,'' in \emph{Advances in Neural Information Processing
  Systems}, 2019, pp. 6007--6018.

\bibitem{bartels1972algorithm}
R.~H. Bartels and G.~W. Stewart, ``Solution of the matrix equation ax+xb=c
  {[F4]} (algorithm 432),'' \emph{Communications of the ACM}, vol.~15, no.~9,
  pp. 820--826, 1972.

\bibitem{golub1979hessenberg}
G.~Golub, S.~Nash, and C.~Van~Loan, ``A hessenberg-schur method for the problem
  ax+ xb= c,'' \emph{IEEE Transactions on Automatic Control}, vol.~24, no.~6,
  pp. 909--913, 1979.

\bibitem{anderson1999lapack}
E.~Anderson, Z.~Bai, C.~Bischof, S.~Blackford, J.~Dongarra, J.~Du~Croz,
  A.~Greenbaum, S.~Hammarling, A.~McKenney, and D.~Sorensen, \emph{LAPACK
  Users' guide}.\hskip 1em plus 0.5em minus 0.4em\relax Siam, 1999, vol.~9.

\bibitem{boyd2011distributed}
S.~Boyd, N.~Parikh, E.~Chu, B.~Peleato, J.~Eckstein \emph{et~al.},
  ``Distributed optimization and statistical learning via the alternating
  direction method of multipliers,'' \emph{Foundations and Trends in Machine
  Learning}, vol.~3, no.~1, pp. 1--122, 2011.

\bibitem{oono2020graph}
K.~Oono and T.~Suzuki, ``Graph neural networks exponentially lose expressive
  power for node classification,'' in \emph{International Conference on
  Learning Representations}, 2020.

\bibitem{xu2018representation}
K.~Xu, C.~Li, Y.~Tian, T.~Sonobe, K.-i. Kawarabayashi, and S.~Jegelka,
  ``Representation learning on graphs with jumping knowledge networks,'' in
  \emph{Proceedings of the 35th International Conference on Machine Learning},
  vol.~80, 2018, pp. 5453--5462.

\bibitem{klicpera2019diffusion}
J.~Klicpera, S.~Wei{\ss}enberger, and S.~G{\"u}nnemann, ``Diffusion improves
  graph learning,'' in \emph{Advances in Neural Information Processing
  Systems}, 2019, pp. 13\,333--13\,345.

\bibitem{sen2008collective}
P.~Sen, G.~Namata, M.~Bilgic, L.~Getoor, B.~Galligher, and T.~Eliassi-Rad,
  ``Collective classification in network data,'' \emph{AI Magazine}, vol.~29,
  no.~3, pp. 93--93, 2008.

\bibitem{nr-aaai15}
R.~Rossi and N.~Ahmed, ``The network data repository with interactive graph
  analytics and visualization,'' in \emph{Proceedings of the Twenty-Ninth
  {AAAI} Conference on Artificial Intelligence}, 2015, pp. 4292--4293.

\bibitem{namata2012query}
G.~M. Namata, B.~London, L.~Getoor, and B.~Huang, ``Query-driven active
  surveying for collective classification,'' in \emph{10th International
  Workshop on Mining and Learning with Graphs}, 2012.

\bibitem{pang2015optimal}
J.~Pang, G.~Cheung, A.~Ortega, and O.~C. Au, ``Optimal graph laplacian
  regularization for natural image denoising,'' in \emph{IEEE International
  Conference on Acoustics, Speech and Signal Processing}, 2015, pp. 2294--2298.

\bibitem{mcauley2015image}
J.~McAuley, C.~Targett, Q.~Shi, and A.~Van Den~Hengel, ``Image-based
  recommendations on styles and substitutes,'' in \emph{Proceedings of the 38th
  International ACM SIGIR Conference on Research and Development in Information
  Retrieval}, 2015, pp. 43--52.

\bibitem{shchur2018pitfalls}
O.~Shchur, M.~Mumme, A.~Bojchevski, and S.~G{\"u}nnemann, ``Pitfalls of graph
  neural network evaluation,'' \emph{arXiv preprint arXiv:1811.05868}, 2018.

\bibitem{kipf2016variational}
T.~N. Kipf and M.~Welling, ``Variational graph auto-encoders,'' \emph{arXiv
  preprint arXiv:1611.07308}, 2016.

\bibitem{li2017diffusion}
Y.~Li, R.~Yu, C.~Shahabi, and Y.~Liu, ``Diffusion convolutional recurrent
  neural network: Data-driven traffic forecasting,'' in \emph{International
  Conference on Learning Representations}, 2018.

\end{thebibliography}

%\begin{IEEEbiography}[{\includegraphics[width=1in,height=1.25in,clip,keepaspectratio]{figs/photo/zhixian_.jpg}}]{Zhixian Chen}
is currently a Ph.D. candidate at the Hong Kong University of Science and Technology (HKUST). She received her BS degree in Mathematics and Applied Mathematics from Hunan University in 2019. Her research generally focuses on developing graph neural networks and graph-based algorithms and applications. 

\end{document}